\documentclass[a4paper,fleqn]{cas-sc}

\usepackage[numbers]{natbib}

\usepackage{amsmath} 
\DeclareMathOperator{\Tr}{Tr}
\usepackage{multirow}
\usepackage{booktabs}
\usepackage{pifont}
\usepackage{xcolor}
\usepackage{float}
\usepackage[caption=false,font=footnotesize]{subfig}
\usepackage{caption}
\newcommand{\cmark}{\ding{51}}%
\newcommand{\xmark}{\ding{55}}%
\usepackage{wasysym}
\usepackage{siunitx}
\usepackage{xurl}

\def\tsc#1{\csdef{#1}{\textsc{\lowercase{#1}}\xspace}}
\tsc{InEKF}

\begin{document}
\let\WriteBookmarks\relax
\def\floatpagepagefraction{1}
\def\textpagefraction{.001}
\shorttitle{Residual-Based Adaptive Kalman Filtering for Legged Robot State Estimation}
\shortauthors{Popescu et~al.}

\title [mode = title]{Residual-Based Adaptive Kalman Filtering for Legged Robot State Estimation}                      

\author[1]{Mihaela Popescu}[orcid=0000-0002-1148-2215]
\cormark[1]
\ead{popescu@uni-bremen.de}
\credit{Conceptualization, Data curation, Formal analysis, Investigation, Methodology, Software, Validation, Visualization, Writing -- original draft}

\author[2]{Dennis Mronga}[orcid=0000-0002-8457-1278]
\credit{Conceptualization, Project administration, Writing -- review and editing}

\author[2,3]{Shivesh Kumar}[orcid=0000-0002-6254-3882]
\credit{Conceptualization, Writing -- review and editing}

\author[1,2]{Frank Kirchner}[orcid=0000-0002-1713-9784]
\credit{Funding acquisition, Supervision}

\affiliation[1]{organization={Robotics Research Group, Faculty of Mathematics and Computer Science, University of Bremen}, 
                city={Bremen},
                citysep={},
                postcode={28359},
                country={Germany}}

\affiliation[2]{organization={Robotics Innovation Center, German Research Center for Artificial Intelligence (DFKI GmbH)},               
                city={Bremen},
                citysep={},
                postcode={28359},
                country={Germany}}

\affiliation[3]{organization={Division of Dynamics, Department of Mechanics \& Maritime Sciences, Chalmers University of Technology},                
                city={Gothenburg},               
                country={Sweden}}

\cortext[cor1]{Corresponding author}

\begin{abstract}
State estimation is a key component in model-based control of walking robots and, more broadly, applicable wherever hidden variables must be inferred. The Kalman filter is widely used to estimate floating-base position and velocity by fusing multiple sensing modalities. However, tuning noise parameters is challenging and typically requires expert knowledge. Moreover, fixed noise parameters are unsuitable for varying gaits and environments. We propose an online adaptation strategy for the process noise covariance matrix $\mathbf{Q}$ and the measurement noise covariance matrix $\mathbf{R}$. Specifically, we introduce a filter residual and innovation-based covariance adaptation method for legged robot state estimation and evaluate it against a baseline approach relying on IMU and foot force measurements. The proposed adaptation is implemented within an Invariant Extended Kalman Filter (InEKF) fusing IMU and leg kinematics. Experiments on indoor and outdoor datasets with a Unitree Go2 quadruped show that adapting $\mathbf{R}$ is sufficient and improves accuracy by 25\% for the trotting gait compared to the fixed-tuned InEKF. Finally, the proposed residual-based adaptation achieves comparable performance to the foot force approach, without requiring foot force measurements or additional parameter tuning.
\end{abstract}

\begin{graphicalabstract}
\includegraphics[width=\linewidth]{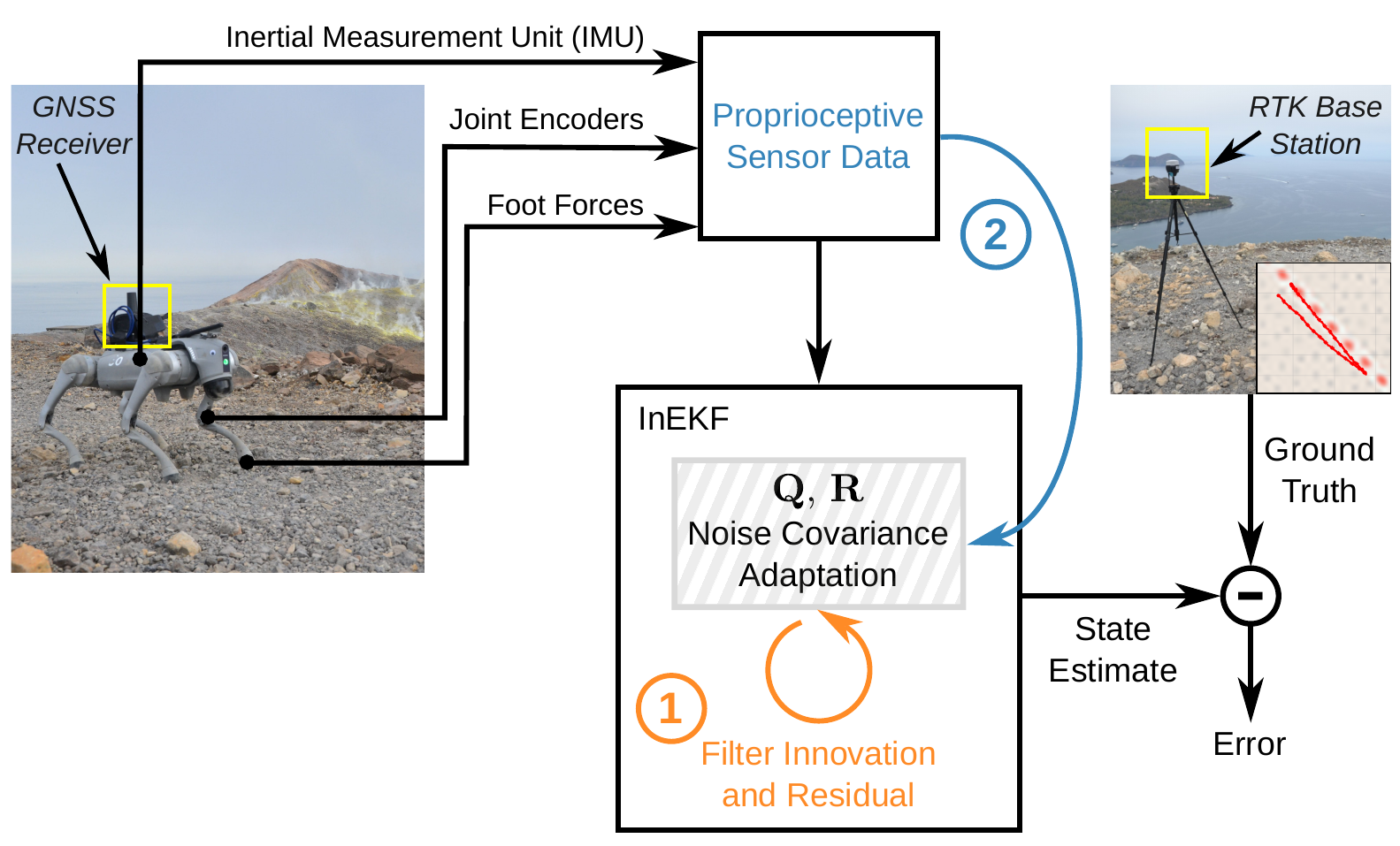}
\end{graphicalabstract}

\begin{highlights}
	\item Performed proprioceptive quadruped state estimation using an Invariant Extended Kalman Filter (InEKF) that fuses inertial measurement unit (IMU) and leg kinematics data.
	\item Proposed the online adaptation of process and measurement noise covariances $\mathbf{Q}$ and $\mathbf{R}$ using filter innovation and residual, benchmarked against an approach based on IMU and foot force measurements.
	\item Improved state estimation accuracy by 15\% overall and 25\% during trotting using online $\mathbf{R}$ covariance adaptation, with the proposed residual-based method requiring no additional sensor data or manual tuning.
	\item Released a dataset for quadruped state estimation covering indoor and outdoor locomotion over 1200 m.
\end{highlights}

\begin{keywords}
	State estimation \sep Legged robots \sep Kalman filter (KF) \sep Sensor fusion \sep Adaptive noise covariance \sep Inertial Measurement Unit (IMU) \sep Leg odometry 
\end{keywords}

\maketitle

\section{Introduction}

Legged robots, such as quadrupeds and humanoids, are highly dynamic systems that must continuously adapt to and operate within unstructured, human-centered environments \cite{bledt2018}. Their ability to traverse complex terrain makes them well suited for hazardous and time-consuming applications, such as search-and-rescue missions and industrial inspection tasks. As the operating environments become increasingly challenging, legged robots require robust and adaptable control frameworks to achieve stable locomotion \cite{cleac2024}. Model-based control approaches rely on accurate, high-rate estimates of the floating-base state to achieve efficient closed-loop control. However, proprioceptive state estimation remains challenging due to several sources of uncertainty, including inertial measurement unit (IMU) drift, as well as leg kinematic errors arising from foot slippage and compliant or uncertain ground contact \cite{jenelten2019}. Kalman filter (KF)-based estimators are widely used to fuse information from multiple sensors and improve state estimation accuracy. However, these estimators typically assume fixed process and measurement noise characteristics, making them unable to adapt to time-varying disturbances affecting sensor measurements, such as foot slippage \cite{bloesch2013b} and impact-induced disturbances in IMU measurements \cite{Driessen2021, Allione2023}. These perturbations typically occur during foot-ground impacts in legged locomotion, with their magnitude and frequency depending on the terrain and gait pattern.

\begin{figure}[]
	\centering
	\includegraphics[width=.9\linewidth]{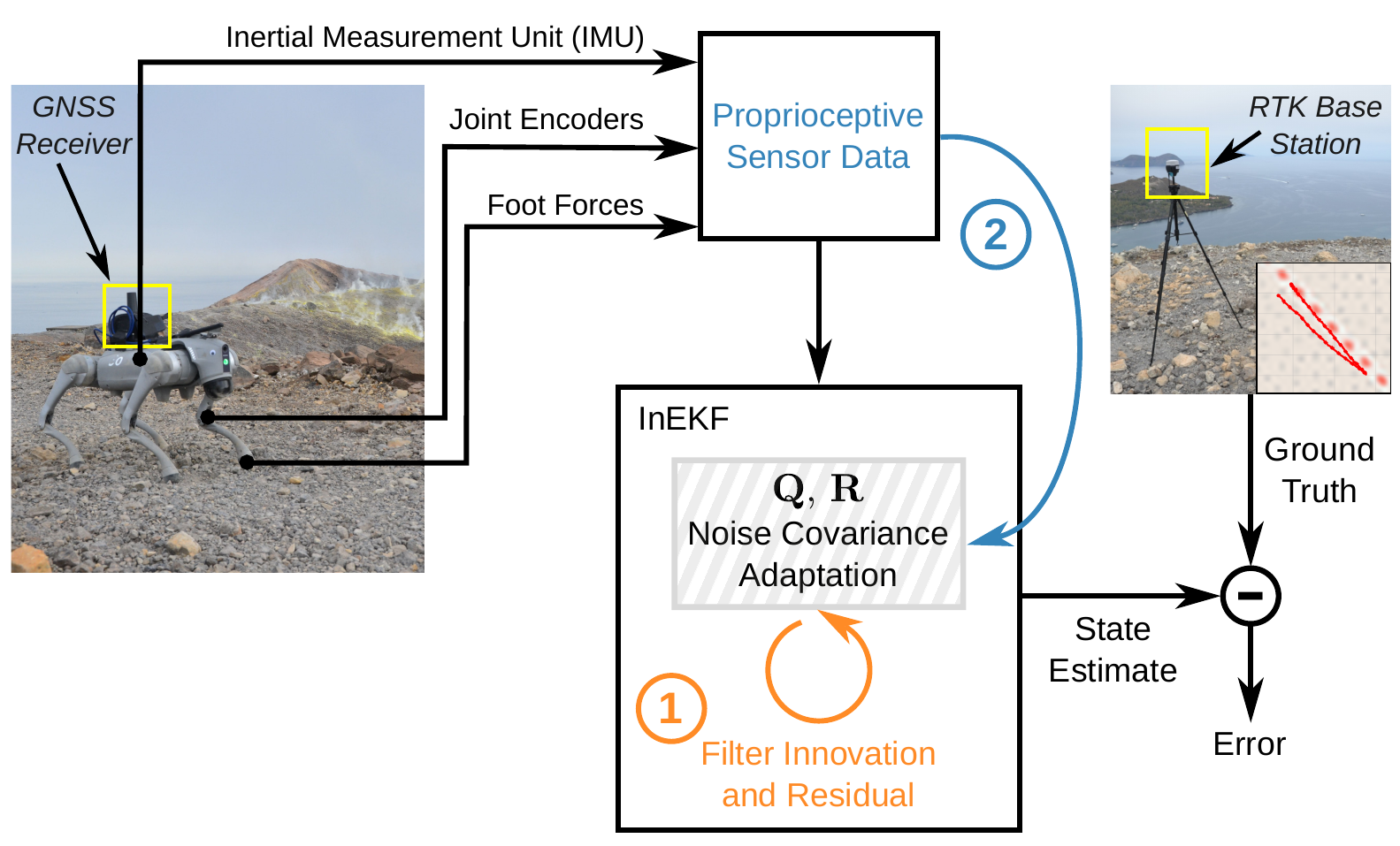}
	\caption{Quadruped state estimation with covariance matrix adaptation based on the proposed (1) filter innovation and residual approach, compared with (2) IMU and foot force sensor data. The methods are evaluated against a fixed-tuned InEKF and validated using ground-truth data from motion capture indoors and accurate GNSS/RTK outdoors.}
	\label{fig:vulcano-cover}
\end{figure}

In this work, we propose an online covariance noise adaptation method for the Invariant Extended Kalman Filter (InEKF) \cite{hartley2020} to cope with variable noise caused by the repetitive making and breaking of leg contact with the ground. Specifically, the process noise covariance $\mathbf{Q}$ associated with the IMU sensor data and the measurement noise covariance $\mathbf{R}$ for leg kinematics are adapted at every iteration. Our approach is based on an innovation and residual-based covariance adaptation technique for Kalman filtering, as described in \cite{mohamed1999}. Specifically, we apply and extend this method within an InEKF framework for legged robot state estimation, which has not previously been explored in this form. Furthermore, we conduct a systematic comparison with a foot force-based covariance adaptation approach from~\cite{camurri2017}. An overview of the proposed methodology is shown in \autoref{fig:vulcano-cover}.
 
We evaluate the methods on real-world data obtained from a Unitree Go2 quadruped robot performing trotting and bio-inspired adaptive gaits in different indoor and outdoor environments. We adapt either $\mathbf{Q}$, $\mathbf{R}$, or both matrices simultaneously, and show that adapting $\mathbf{R}$ covariance alone is sufficient for our use case. Moreover, we demonstrate that $\mathbf{R}$ covariance matrix adaptation improves state estimation performance by up to 25\% on the trotting gait and 15\% overall. In terms of performance, the proposed method based on KF innovation and residual is comparable to the foot force approach, with the advantage of not requiring hyperparameter tuning or additional sensor data, which makes it more suitable for deployment on real-world scenarios. Finally, we release a quadruped dataset\footnote{Quadruped dataset available in Zenodo: \url{https://zenodo.org/records/19336009}} recorded both indoors, with accurate ground truth from a motion capture system, and outdoors on Vulcano Island, with GPS ground truth.

In summary, the main contributions of this work are:

\begin{enumerate}[i]	
	\item Proposed the novel application and extension of an online noise adaptation technique for legged robot state estimation based on Kalman filter innovation and residual for IMU and leg kinematics. 
	
	\item Conducted an ablation study of $\mathbf{Q}$ and $\mathbf{R}$ adaptation, showing that $\mathbf{R}$ measurement noise adaptation alone corresponding to leg odometry is sufficient to improve estimation performance.
	
	\item Performed extensive real-world validation on a quadruped robot across flat indoor and unstructured outdoor terrains over a distance of 1200 m, and released the dataset open source.
	
	\item Improved state estimation accuracy by an average of 15\% overall and up to 25\% during trotting, compared to a fixed-tuned InEKF.
\end{enumerate}

The paper is organized as follows. Section~\ref{sec:related-work} presents the state of the art on KF adaptation, and Section~\ref{sec:state-estimation} introduces the fundamentals of InEKF, which is used in this work. Section~\ref{sec:covariance-adaptation} describes the methods used for covariance matrix adaptation, while Section~\ref{sec:experimental-setup} introduces the experimental setup and the evaluation datasets. The results are discussed in Section~\ref{sec:results}, followed by a discussion in Section~\ref{sec:discussion}. Finally, the conclusion and outlook are presented in Section~\ref{sec:conclusion}.

\section{Related Work}
\label{sec:related-work}

This section presents various state-of-the-art approaches for handling contact uncertainty in legged robot state estimation, as well as innovation and residual-based adaptive Kalman filtering methods for robotic applications.

\subsection{Contact Uncertainty in Legged Robot State Estimation}
\label{sec:sota-quad-state-est}

Proprioceptive state estimation for legged robots using IMU and leg kinematics enables efficient, high-frequency estimation required for closed-loop feedback control \cite{bloesch2013}. However, leg odometry is often affected by issues such as link flexibilities \cite{Romualdi2022}, uncertain contacts and foot slippage, for which various mitigation strategies have been proposed.

In terms of robustness to outliers, \cite{bloesch2013b} enhances the Unscented Kalman Filter (UKF) by rejecting unreliable leg kinematic measurements on unstable and slippery terrain. The approach discards uncertain measurements based on the Mahalanobis distance of the innovation, requiring manual tuning of the rejection threshold. Similarly, \cite{Kim2021} proposes a Maximum A Posteriori (MAP)-based approach for quadruped state estimation and slip rejection, where a foot in contact with the ground is classified as slipping when the L2 norm of its estimated velocity exceeds a tunable threshold. Building on this, \cite{Yoon2024} assigns a fixed, manually tuned high covariance to slipping measurements instead of discarding them. While these approaches improve robustness through explicit slip handling, they treat noisy leg kinematics in a binary manner and are therefore less flexible in handling partially reliable measurements. More recently, \cite{santana2024} improves robustness to outliers by replacing the L2 norm with robust cost functions for leg measurement updates in the InEKF, while keeping the sensor noise covariance matrix fixed.

Furthermore, to improve contact detection, \cite{bledt2018} proposed a probabilistic approach that fuses the planned gait phase with measured foot height and contact forces to estimate the contact probability, which is then converted into a binary contact state. However, the noise parameters of the leg kinematic measurements remain unchanged. In contrast, \cite{camurri2017} presents a probabilistic contact estimation framework that scales leg kinematics measurements and introduces an online adaptation of the measurement noise covariance by combining the consistency of each leg's contribution and impact forces, weighted by scale factors. 
This approach has also been extended to error-state Kalman filters \cite{kang2023} and the InEKF \cite{wan2024}, enabling quadruped locomotion on soft, uneven, and non-stationary terrain. However, accurate foot force measurements are required, and the scale factors need to be manually tuned.

Learning-based approaches have also been proposed to improve legged robot state estimation. For instance, \cite{yang2024} includes a learning-based method to weight leg contributions in the InEKF. Namely, a neural network is trained with foot force data and leg kinematics to produce a fused leg kinematics measurement. More recently, \cite{baumgartner2026} proposed CoCo-InEKF, a differentiable InEKF that learns continuous contact velocity covariances instead of relying on binary contact states, demonstrated on a small-size biped robot. A lightweight neural network was trained end-to-end using a state estimation loss, removing the need for heuristic ground-truth contact labels. However, both approaches rely on learned models whose performance and generalization may not readily transfer across different robot platforms.
 
\subsection{Adaptive Kalman Filtering based on Innovation and Residual}
\label{sec:sota-adaptive-kf}

More broadly, several works address the online adaptation of noise parameters for KF-based position and orientation tracking. The work in \cite{mohamed1999} proposes an adaptive Kalman filter for integrated inertial navigation and global positioning system (INS/GPS) state estimation. The method uses filter innovation and residual to adapt the process noise covariance $\mathbf{Q}$, the measurement noise covariance $\mathbf{R}$, or both simultaneously, without requiring additional parameter tuning. Building on this work, \cite{ding2007} derives a scaled adaptation algorithm for $\mathbf{Q}$ assuming a known measurement noise covariance $\mathbf{R}$, while \cite{akhlaghi2017} introduces a forgetting factor into the innovation and residual computation for rotor angle and speed estimation. More recently, \cite{kim2025} applied innovation-based covariance adaptation over a moving window to quadruped state estimation by adapting the foot-slip component of $\mathbf{Q}$. However, the potential of using this approach to adapt the measurement noise covariance $\mathbf{R}$ in legged robot state estimation remains largely unexplored.

Finally, the studies in \cite{or2023, cohen2025} rely on the approach from \cite{mohamed1999} to learn the noise matrix $\mathbf{Q}$ for inertial sensing. The IMU data is then fused with Doppler velocity log (DVL) to estimate the state of an autonomous underwater vehicle. However, learning methods have drawbacks such as increased computational demand, lack of explainability and poor generalization. Hence, we use classical adaptive Kalman filter methods to ensure speed, stability and interpretability for legged robot state estimation. 

\section{State Estimation Framework}
\label{sec:state-estimation}

The InEKF \cite{hartley2020} is employed for state estimation due to its strong convergence properties and reduced sensitivity to linearization errors. The left-invariant, world-centric formulation is adopted, consistent with the world-frame state representation and contact-aided leg kinematics. The state matrix $\mathbf{X}$ is defined on the Lie group $SE_{N+2}(3)$ as follows:
\begin{equation}
	\boldsymbol X = \begin{bmatrix}
		\boldsymbol R_B^W & \boldsymbol v & \boldsymbol p & \boldsymbol p_{c_1} & \ldots &  \boldsymbol p_{c_N} \\
		\boldsymbol0_{1, 3} & 1 & 0 & 0 & \ldots & 0 \\
		\boldsymbol0_{1, 3} & 0 & 1 & 0 & \ldots & 0 \\
		\vdots & \vdots & \vdots & \vdots & \ddots & \vdots \\
		\boldsymbol0_{1, 3} & 0 & 0 & 0 & $\ldots$ & 1 
	\end{bmatrix}
	\label{eq:state_matrix}
\end{equation}
\noindent where $\boldsymbol R_B^W$, $\boldsymbol v$ and $\boldsymbol p$ represent the orientation, velocity and position of the robot's floating base in the world frame, and $\boldsymbol p_{c_i}$ is the position of the feet $i \in N$ in contact with the ground in the world frame. Furthermore, the IMU gyroscope and accelerometer bias $\boldsymbol \theta = [\theta_g \ \theta_a]^T$ is also estimated.

The estimation error $\boldsymbol{\xi}_t$ is defined on the Lie algebra $se_{N+2}(3)$, which corresponds to the first-order linearization of the Lie group $SE_{N+2}(3)$ via the logarithmic map:
\begin{equation}
	\boldsymbol{\xi}_t = \log\!\left(\mathbf{X}_t^{-1}\hat{\mathbf{X}}_t\right)
\end{equation}
where $\mathbf{X}_t$ is the true state and $\hat{\mathbf{X}}_t$ is its estimate. Subsequently, the state error covariance matrix is obtained as $\mathbf{P}_t = \mathbb{E}[\boldsymbol{\xi}_t \boldsymbol{\xi}_t^\top]$, representing the uncertainty of the state estimate on $se_{N+2}(3)$.

\subsection{Prediction}
The filter propagates the estimate on the Lie group and the covariance on the associated Lie algebra. The prediction relies on the input vector $\mathbf{u} = [\boldsymbol{\omega} \ \boldsymbol{a}]^T$, consisting of the robot's true angular velocity $\boldsymbol{\omega}$ and linear acceleration $\boldsymbol{a}$ expressed in the body frame. The IMU sensor measurements are modeled as noisy observations as follows:
\begin{equation}
	\begin{aligned}
		\tilde{\boldsymbol{\omega}} &= \boldsymbol{\omega} + \mathbf{n}_g, \\
		\tilde{\boldsymbol{a}} &= \boldsymbol{a} + \mathbf{n}_a,
	\end{aligned}
\end{equation}
where $\mathbf{n}_g \sim \mathcal{N}(\mathbf{0}, \boldsymbol{\sigma}_g^2)$ and
$\mathbf{n}_a \sim \mathcal{N}(\mathbf{0}, \boldsymbol{\sigma}_a^2)$ represent additive zero-mean Gaussian noise, with $\boldsymbol{\sigma}_{g}$ and $\boldsymbol{\sigma}_{a}$ denoting the standard deviations of the IMU gyroscope and accelerometer measurements along the $\{x, y, z\}$-axes, respectively. The associated process noise covariance $\mathbf{Q}$ for the full state is the diagonal matrix:
\begin{equation}
	\mathbf{Q} = \begin{bmatrix}
		\boldsymbol{\sigma}_g^2 & 0 & 0 & 0\\
		0 & \boldsymbol{\sigma}_a^2 & 0 & 0\\
		0 & 0 & \boldsymbol{\sigma}_{\boldsymbol p_{c}}^2 & 0\\
		0 & 0 & 0 & \boldsymbol{\sigma}_{\boldsymbol \theta}^2\\
	\end{bmatrix}
\end{equation}
where $\boldsymbol{\sigma}_{\boldsymbol p_{c}}$ denotes the contact linear velocity noise due to slippage, and $\boldsymbol{\sigma}_{\boldsymbol \theta}$ denotes the IMU bias standard deviation. The discrete-time state propagation is obtained using forward Euler integration, with orientation updated on the Lie group $SO(3)$, and is given by:
\begin{equation}
	\begin{aligned}
		\mathbf{R}_{t+1} &= \mathbf{R}_{t} \exp\!\left( \left(\tilde{\boldsymbol{\omega}}_{t} - \boldsymbol{\theta}_{g,t} - \mathbf{n}_{g,t}\right)\Delta t \right), \\
		\mathbf{v}_{t+1} &= \mathbf{v}_{t} + \left( \mathbf{g} + \mathbf{R}_{t}\left(\tilde{\boldsymbol{a}}_{t} - \boldsymbol{\theta}_{a,t} - \mathbf{n}_{a,t}\right) \right)\Delta t, \\
		\mathbf{p}_{t+1} &= \mathbf{p}_{t} + \mathbf{v}_{t}\Delta t,
	\end{aligned}
	\label{eq:imu_discrete_integration}
\end{equation}

The predicted covariance $\mathbf{P}_t^-$ is obtained as follows:
\begin{equation}
	\mathbf{P}_t^- =
	\mathbf{A}_t \mathbf{P}_{t-1}^+ \mathbf{A}_t^\top + \mathbf{G}_t \mathbf{Q} \mathbf{G}_t^\top,
	\label{eq:cov_propagation}
\end{equation}
where $\mathbf{A}_t$ denotes the linearized state transition matrix and $\mathbf{G}_t$ is the process noise Jacobian.

\subsection{Update}
The update step of the InEKF uses leg kinematics obtained from joint encoder measurements when foot $i \in N$ is in contact with the ground. The foot position $\tilde{\mathbf{p}}_{c_i}^{B}$ expressed in body frame is modeled as follows:
\begin{equation}
	\tilde{\mathbf{p}}_{c_i}^{B}
	=
	(\boldsymbol R_B^W)^{\top}
	\left(
	\boldsymbol p_{c_i}-\mathbf{p}
	\right)
	+
	\mathbf{n}_{l},
	\qquad
	\mathbf{n}_{l}\sim\mathcal{N}(\mathbf{0},\mathbf{R}),
	\label{eq:encoder_measurement}
\end{equation}
where the measurement noise covariance matrix $\mathbf{R}$ is a diagonal matrix with elements $\boldsymbol{\sigma}_{l}^2$, representing the variances of the leg kinematics measurements along the $\{x, y, z\}$-axes:
\begin{equation}
	\mathbf{R} = \begin{bmatrix}
		\sigma_{l_x}^2 & 0 & 0\\
		0 & \sigma_{l_y}^2  & 0 \\
		0 & 0  & \sigma_{l_z}^2 \\
	\end{bmatrix}
\end{equation}

Unlike the original InEKF formulation in \cite{hartley2020}, where the measurement covariance is obtained by propagating the joint encoder uncertainties through the forward kinematics Jacobian, this work directly models the uncertainty of the foot position measurement using a diagonal covariance matrix. This simplification allows the measurement noise to be tuned independently along each Cartesian axis while retaining the same update formulation. 

Finally, the Kalman gain used to fuse the IMU prediction with the leg kinematics update is  derived from the state and measurement noise covariance matrices $\mathbf{P}$ and $\mathbf{R}$ as follows:
\begin{equation}
	\mathbf{K}_{t} =
	\mathbf{P}_t^- \mathbf{H}_{t}^\top
	\left(
	\mathbf{H}_{t} \mathbf{P}_t^- \mathbf{H}_{t}^\top + \mathbf{R}_{t}
	\right)^{-1},
	\label{eq:kalman_gain}
\end{equation}
where $\mathbf{H}_{t}$ is the measurement Jacobian associated with the leg kinematics, and $\mathbf{P}_t^-$ depends on $\mathbf{Q}$ as shown in \autoref{eq:cov_propagation}. 

\section{Covariance Matrix Adaptation}
\label{sec:covariance-adaptation}

The Kalman gain, which is responsible for sensor fusion in the Kalman filter, is fundamentally determined by the noise covariance matrices $\mathbf{Q}$ and $\mathbf{R}$. Under ideal conditions, adjusting only one of these matrices is theoretically sufficient, since the Kalman gain primarily depends on their relative weighting. However, this assumption may not hold in practice due to factors such as model mismatch, linearization errors, and unmodeled disturbances. Therefore, we adapt either $\mathbf{Q}$, $\mathbf{R}$, or both simultaneously to assess whether simultaneous adaptation provides additional benefits under real-world conditions and to determine which covariance matrix has the greatest impact on estimation performance.

For this, we propose a novel method to adapt $\mathbf{Q}$ and $\mathbf{R}$ based on (i)~filter innovation and residual and compare it to a baseline approach based on (ii)~IMU and foot force sensor data. To preserve the symmetric positive definite (SPD) property of the covariance matrices and prevent divergence, the adaptation is restricted to the diagonal elements. An overview of the adapted noise parameters is provided in \autoref{table:adaptation}.

\begin{table}[]
	\vspace{2mm}
	\begin{center}			
		\caption{Overview of the $\mathbf Q$ and $\mathbf R$ matrix adaptation for each method. Adapted elements are denoted by {\cmark}, non-adaptable elements by {\xmark}, and adaptable elements that have not been adapted in this work by \LEFTcircle. }
		
		\begin{tabular}{ccccc} 
			\toprule		
			\multirow{4}{*}{\shortstack{Covariance\\Matrix}}  &\multirow{4}{*}{\shortstack{Std.\\ Dev.}}  & \multirow{4}{*}{\shortstack{Description}} & \multicolumn{2}{c}{Adaptation Method} \\
			\cmidrule{4-5} 
			& &  & IMU and Foot Force & Filter Innovation \\
			& &  & Sensor Data & and Residual \\
			\midrule
			\multirow{4}{*}{$\mathbf Q$} 	& $\boldsymbol{\sigma}_g$ & gyroscope 	&	\cmark	&\cmark	\\ 
			&	$\boldsymbol{\sigma}_a$ & accelerometer	& \cmark &	\cmark	\\
			&	$\boldsymbol{\sigma}_{\boldsymbol p_{c}}$ &	foot slip		&	\xmark &	\LEFTcircle	\\ 
			&	$\boldsymbol{\sigma}_{\boldsymbol \theta}$ & IMU bias	&   \xmark & \LEFTcircle \\ 
			\midrule
			$\mathbf R$	& $\boldsymbol{\sigma}_l $ & leg kinematics	&	\cmark	&	\cmark	\\ 
			\bottomrule
		\end{tabular}
		\label{table:adaptation}
	\end{center}
\end{table}

\subsection{Adaptation based on Filter Innovation and Residual}

In this work, we propose leveraging internal Kalman filter quantities, namely the innovation $\boldsymbol \nu_t^-$ and residual $\boldsymbol \nu_t^+$, to adapt the $\boldsymbol{Q}$ and $\boldsymbol{R}$ covariance matrices for legged robot state estimation. This exploits the iterative nature of the Kalman filter, which fuses model predictions with sensor measurements. The innovation and residual are defined as:
\begin{align}
	\boldsymbol \nu_t^- &= \mathbf z_t - \mathbf{H}_{t}\mathbf X_t^-\\	
	\boldsymbol \nu_t^+ &= \mathbf z_t - \mathbf{H}_{t}\mathbf X_t^+
\end{align}
\noindent where $\mathbf z_t$ is the sensor measurement, $\mathbf X_t^-$ is the predicted state, and $\mathbf X_t^+$ is the updated state. Specifically, for legged robot state estimation, the innovation and residual for each contact leg $i$ at time $t$ are defined as follows: 
\begin{equation}
	\boldsymbol{\nu}_{i, t}^{\{-, +\}} = \boldsymbol R_B^W \tilde{\boldsymbol{p}}^B_{c_{i}, t}- (\boldsymbol p_{c_i} - \boldsymbol p)
\end{equation}
where $\boldsymbol R_B^W$, $\boldsymbol p$ and $\boldsymbol p_{c_i}$ correspond to the body orientation, body position and foot position from the filter states $\mathbf X_t^-$ and $\mathbf X_t^+$ for the innovation and residual formulations, respectively. Next, the covariance $\boldsymbol C_{\boldsymbol \nu_{i, t}}^{\{-, +\}}$ of the innovation and residual for each leg $i$ is obtained over a sliding window:
\begin{equation}
	\boldsymbol C_{\boldsymbol \nu_{i, t}^{\{-, +\}}} = \frac{1}{M}\sum_{j=j_0}^t \boldsymbol \nu_{i, j}^{\{-, +\}}\left(\boldsymbol \nu_{i, j}^{\{-, +\}}\right){}^T
\end{equation}
where M is the window size and $j_0 = t-M+1$ is the first element of the estimation window. The adaptation of the $\boldsymbol Q$ and $\boldsymbol R$ covariance matrices is described next.

\subsubsection{Process Noise Covariance Matrix $\boldsymbol Q$} The adaptation is based on the innovation covariance $\boldsymbol C_{\boldsymbol \nu_t^-}$ as described in~\cite{ding2007}. Since the Kalman gain depends on the relative magnitudes of $\boldsymbol P$ and $\boldsymbol R$, the process noise covariance $\boldsymbol Q$ is obtained as a factor of $\boldsymbol R$, assuming that the $\boldsymbol R$ covariance matrix is known. The scaling factor $\alpha_i$ for each leg $i$ is computed as follows:
\begin{equation}
	\alpha_i = \frac{\Tr \left(\boldsymbol C_{\boldsymbol \nu_{i, t}^-} - \boldsymbol R_{i, t}\right)}{\Tr \left(\boldsymbol H_{i, t} \boldsymbol P^-t \boldsymbol H^T{i, t}\right)}
\end{equation}
\noindent where $\Tr(\cdot)$ denotes the matrix trace operator.

The final scaling factor $\alpha$ is obtained as the average of $\alpha_i$ over all legs currently in contact with the ground. To ensure stability at high update rates, we do not recursively scale the previous $\boldsymbol Q_{t-1}$ as proposed in~\cite{ding2007}. Instead, we apply the scaling factor directly to the nominal, fixed-tuned matrix $\boldsymbol Q$:
\begin{equation}
	\boldsymbol Q_t=\boldsymbol Q\sqrt{\alpha}
\end{equation}

It is important to note that the adaptation affects only the IMU uncertainty, namely the gyroscope and accelerometer noise parameters $\boldsymbol{\sigma}_g$ and $\boldsymbol{\sigma}_a$. The averaging over feet in contact is solely used to obtain a robust estimate of the common scaling factor $\alpha$, and should not be interpreted as an adaptation of individual foot measurements or slip characteristics. In contrast to \cite{kim2025}, where the innovation covariance is used to adapt the foot slip component of $\boldsymbol Q$, the proposed approach focuses exclusively on IMU process noise adaptation. Foot slip effects are instead handled separately through adaptation of the measurement noise covariance $\boldsymbol R$ using filter residuals, as described in the following section.

\subsubsection{Measurement Noise Covariance Matrix $\boldsymbol R$}
The proposed $\boldsymbol R$ adaptation is based on the residual covariance $\mathbf{C}_{\boldsymbol{\nu}_t^+}$, following~\cite{mohamed1999}, and is extended here to legged robot state estimation. The residual formulation has the advantage that, unlike the innovation-based alternative, it preserves the SPD property of the covariance matrix $\boldsymbol{R}$.

Since a left-invariant InEKF is used, the state covariance $\boldsymbol{P}$ is expressed in the body frame, while the measurement covariance $\boldsymbol{R}$ is defined in the world frame. To ensure frame consistency, the covariance is rotated using $\boldsymbol{R}_B^W$ from the updated filter state $\mathbf{X}_t^+$, leading to the following adaptation rule:
\begin{equation}
	\boldsymbol R_{i, t} = \boldsymbol C_{\boldsymbol \nu_{i, t}^+} + \boldsymbol R_B^W(\boldsymbol H_{i, t} \boldsymbol P_t^+ \boldsymbol H^T_{i, t}) ( \boldsymbol{R}_B^W )^T
\end{equation}
\noindent where $\boldsymbol R_{i, t}$ is the measurement noise covariance matrix adapted for each leg $i$ in contact with the ground separately.

\subsection{Adaptation based on IMU and Foot Force Data}
\label{sec:propriceptive_method}

\begin{figure}[]
	\centering
	\includegraphics[width=.67\linewidth]{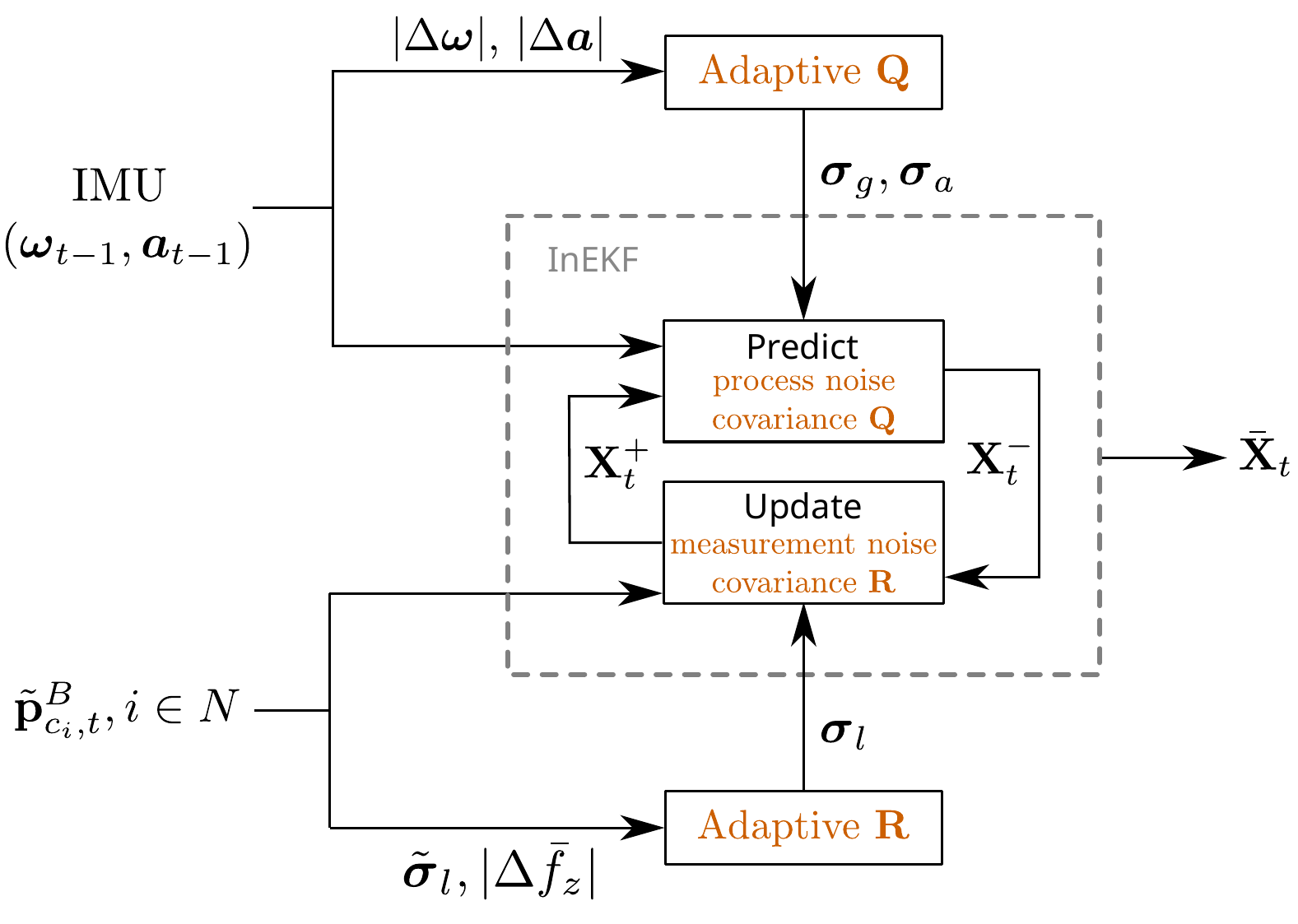}
	\caption{Adaptation of the $\boldsymbol Q$ and $\boldsymbol R$ matrices based on IMU measurements, foot forces and leg kinematics within the InEKF state estimator.}	
	\label{fig:formula_adaptation}
\end{figure}

In the baseline approach, proprioceptive sensor data is used for online adaptation of the noise covariance matrices, such as IMU measurements, leg kinematics, and foot forces. A schematic representation is shown in \autoref{fig:formula_adaptation}.

\subsubsection{Process Noise Covariance Matrix $\boldsymbol Q$}
During locomotion, ground contact events can introduce transient disturbances in the IMU measurements, which temporarily degrade attitude estimation performance \cite{Driessen2021, Allione2023}. To account for these time-varying uncertainties, we introduce an IMU-driven adaptation of the process noise covariance $\boldsymbol Q$. In particular, variations in consecutive gyroscope and accelerometer readings are used as indicators of contact-induced excitation, and are employed to increase the IMU noise parameters $\boldsymbol{\sigma}_g$ and $\boldsymbol{\sigma}_a$ during impact events as follows:
\begin{align}
	\boldsymbol \sigma_{g} &= \sqrt{\boldsymbol \sigma_{g_0}^2 + \left(\boldsymbol \sigma_{g_0}{\alpha_1} \left|\Delta{\boldsymbol{\omega}}\right| \right)^2}	\\
	\boldsymbol \sigma_{a} &= \sqrt{\boldsymbol \sigma_{a_0}^2 + \left(\boldsymbol \sigma_{a_0}{\alpha_1} \left|\Delta{\boldsymbol{a}}\right| \right)^2}
\end{align}
\noindent where $\alpha_1$ is a configurable gain, $\boldsymbol\sigma_{g_0}$ and $\boldsymbol\sigma_{a_0}$ denote the initial gyroscope and accelerometer standard deviations, and $\left|\Delta{\boldsymbol{\omega}}\right|$ and $\left|\Delta{\boldsymbol{a}}\right|$ represent the absolute temporal differences of IMU angular velocity $\tilde{\boldsymbol{\omega}}$ and linear acceleration $\tilde{\boldsymbol{a}}$:
\begin{align}
	\left|\Delta{\boldsymbol{\omega}}\right| &= \left|\tilde{\boldsymbol{\omega}}_{t} - \tilde{\boldsymbol{\omega}}_{t-1}\right|
	\label{eq:delta_w} \\
	\left|\Delta{\boldsymbol{a}}\right| &= \left|\tilde{\boldsymbol{a}}_{t} - \tilde{\boldsymbol{a}}_{t-1}\right|
	\label{eq:delta_a}
\end{align}

\subsubsection{Measurement Noise Covariance Matrix $\boldsymbol R$} Based on \cite{camurri2017}, the baseline adaptation for each leg $i \in N$ in contact with the ground relies on the consistency of leg kinematic updates and foot force difference as follows:
\begin{align}
	\boldsymbol\sigma_{l, i} &= \sqrt{\boldsymbol\sigma_{l_0}^2 + \left({\alpha_2} \tilde{\boldsymbol\sigma}_{l} + \left(1-{\alpha_2}\right)\boldsymbol\sigma_{l_0}{\alpha_3}\left|f_{z_i, t} - f_{z_i, t-1} \right|\right)^2}
\end{align}
\noindent where ${\alpha_2}, {\alpha_3}$ are tunable parameters, $\boldsymbol\sigma_{l_0}$ is the initial leg kinematics standard deviation, $\tilde{\boldsymbol\sigma}_{l}$ is the standard deviation of the leg odometry measurements $\tilde{\mathbf{p}}^B_{c_{i}}$ for the legs $i \in N$, and $ f_{z_i, t}$ is the foot force of leg $i$ on the $z$-axis at time $t$. In contrast to \cite{camurri2017}, we obtain the noise parameters for each leg individually for a more precise adaptation, rather than averaging the foot force differences across all legs in contact.

\section{Experimental Setup}
\label{sec:experimental-setup}

\subsection{Robotic Platform}

For real-world experiments, we used a Unitree Go2 quadruped robot equipped with a 6-DOF IMU sensor with gyroscope and accelerometer, as well as joint encoders for leg kinematics and foot force / torque (F/T) sensors for contact detection. The control architecture consists of a gait sequencer for planning footsteps with different gait types, a model predictive controller (MPC) for planning contact forces, and a whole-body controller (WBC) for executing the desired motions using full-body dynamics, as described in \cite{stark2025}.

\subsection{Datasets}
For evaluation purposes, we collected a dataset with the quadruped robot in indoor and outdoor environments, as shown in \autoref{table:datasets}, and released the data open-source via Zenodo~\cite{popescu2026}. A video with sample trajectories is attached.

\begin{table}[] 
	\vspace{2mm}
	\begin{center}			
		\caption{Overview of collected dataset with different trajectories and gaits. The indoor dataset recorded in the lab is denoted by $\text{L}$, and the outdoor dataset from the Vulcano field trials by $\text{F}$. The distance covered by each trajectory is marked by $\text{D}$.}
			\begin{tabular}{rcccr} 
				\toprule				
				{\centering Index} & Gait & Terrain & Trajectory & D [m] \\
				\midrule
				L1	&	trotting	&	flat	&	random	&	22.0	\\ 			
				L2	&	trotting	&	stones	&	random	&	31.1	\\
				L3	&	trotting	&	flat	&	turning counterclockwise	&	51.4\\ 			
				L4	&	trotting	&	flat	&	turning clockwise	&	64.7	\\
				L5	&	adaptive	&	flat	&	random	&	26.4	\\
				L6	&	adaptive	&	stones	&	random	&	35.5	\\
				L7	&	trotting	&	flat	&	straight line, long	&	175.2	\\ 			
				L8	&	trotting	&	flat	&	turning, long	&	104.5	\\
				L9	&	adaptive	&	flat	&	straight line, long	&	167.2	\\
				\midrule
				F1	&	trotting	&	rocky	&	straight line	&	10.8	\\
				F2	&	trotting	&	rocky	&	straight line	&	11.7	\\
				F3	&	adaptive trot	&	rocky	&	straight and back	&	29.2	\\
				F4	&	trotting	&	rocky	&	straight line	&	17.8	\\
				F5	&	trotting	&	rocky	&	straight line	&	9.8	\\
				F6	&	adaptive trot	&	rocky	&	straight one way, curves back	&	28.6\\
				F7	&	trotting		&	rocky	&   straight one way, loop back	&	29.6	\\
				F8	&	adaptive walk	&	rocky	&	straight one way, curves back	&	27.0	\\
				F9	&	adaptive trot	&	rocky	&	straight and back	&	13.3	\\
				F10	&	trotting, adaptive trot	&	rocky	&	step in place, straight and back 	&	18.1	\\
				F11	&	trotting, adaptive trot	&	rocky	&	step in place, straight and back	&	19.1	\\
				F12	&	adaptive trot   &	rocky	&		straight line	&	9.8	\\
				F13	&	trotting, adaptive trot  &	rocky	&	step in place, straight and back	&	20.5	\\
				F14	&	trotting, adaptive trot  &	rocky	&		straight and back	&	14.1\\
				F15	&	adaptive trot	&	rocky	&		straight and back	&	123.1	\\
				F16	&	trotting &	rocky	&		straight line	&	25.3	\\
				F17	&	adaptive trot	&	rocky	&		straight line	&	95.8	\\
				F18	&	adaptive trot	&	rocky	&			straight line	&	25.0	\\
				\bottomrule
		\end{tabular}
		\label{table:datasets}
	\end{center}
\end{table}

\subsubsection{Indoor (Lab)} 
\begin{figure}[]
	\centering	
	\subfloat[Top view of the Unitree Go2 quadruped with reflective markers for Vicon tracking and coordinate frame definition.]{
		\includegraphics[width=0.47\columnwidth]{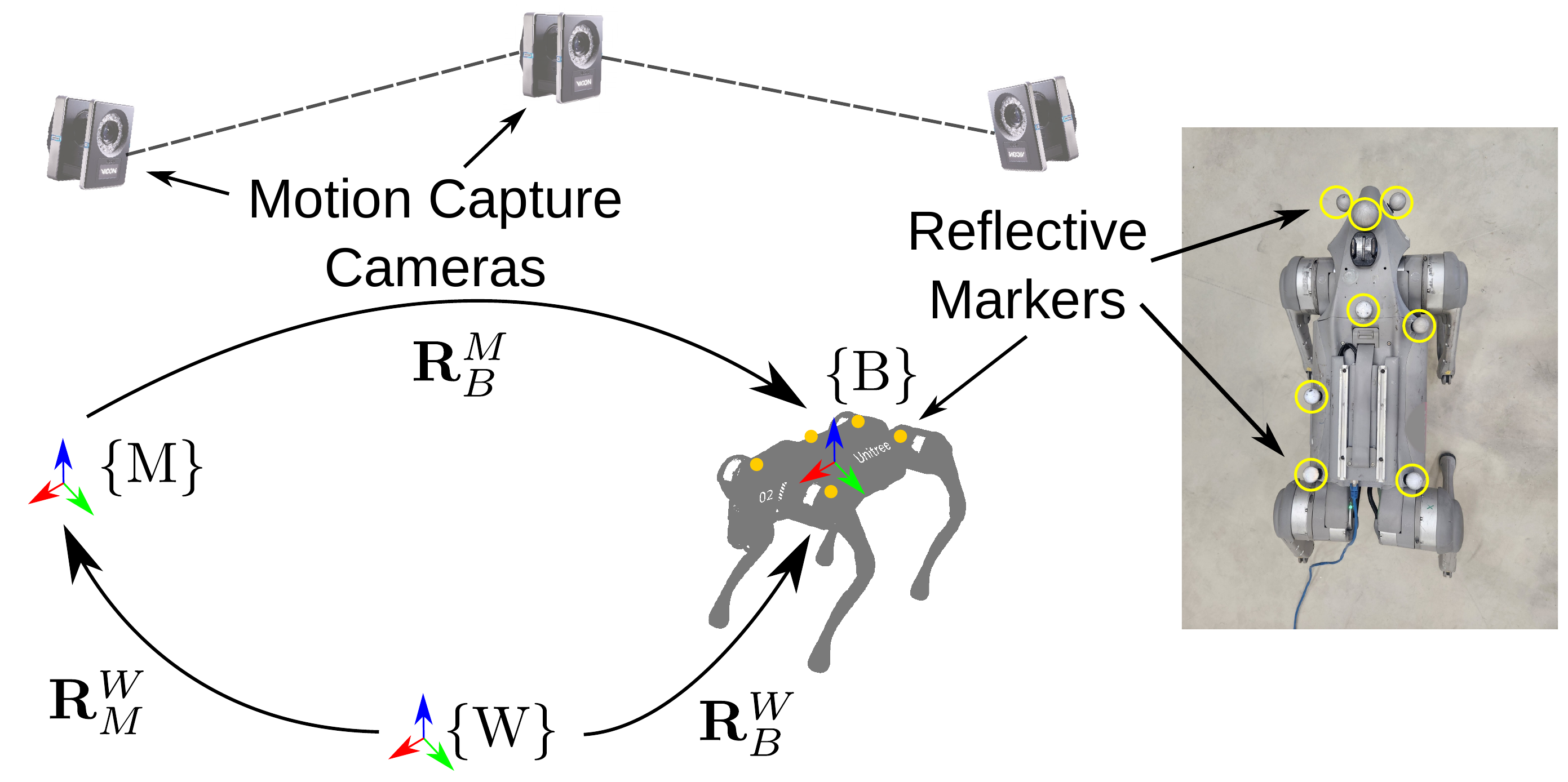}
		\label{fig:coord_frames}
	}
	\hfill
	\subfloat[Walking experiments were performed indoors and outdoors on Vulcano Island, Italy, with the Unitree Go2 quadruped robot.]{
		\includegraphics[width=0.47\columnwidth]{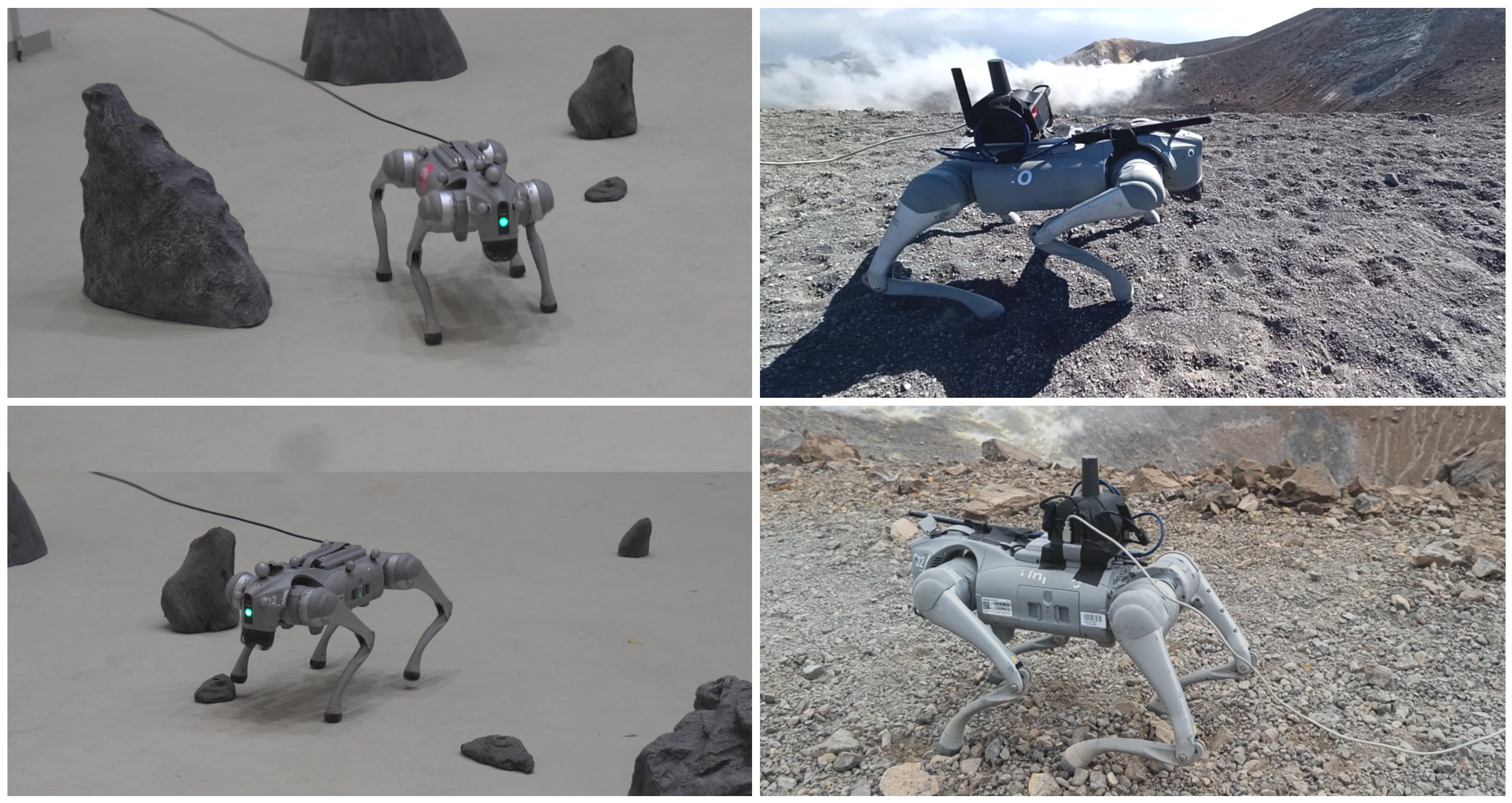}
		\label{fig:dataset}
	}
	
	\caption{Experimental setup and datasets used in this work.}
	\label{fig:setup_and_dataset}
\end{figure}

The indoor dataset consists of 9 trajectories performed with two gaits: trotting and an adaptive, bio-inspired approach that switches between walking and trotting based on the commanded velocity, as presented in \cite{stark2025}. The motions are executed on a flat floor with occasional stones on the way. The trajectories cover a distance of 680 m, with both IMU and joint encoders recorded at 1000 Hz. Ground truth was obtained via a Vicon motion capture system at 100 Hz using reflective markers as shown in Figure~\ref{fig:coord_frames}. We denote the robot body frame as \{B\}, the robot world frame as \{W\} and the motion capture world frame as \{M\}. To align the state estimation with the ground truth, we used the initial robot yaw orientation measured by Vicon. Since both coordinate systems \{W\} and \{M\} are initialized on the same plane, roll and pitch alignment was omitted to avoid inaccuracies caused by marker placement and rigid body definition.

\subsubsection{Outdoor (Field)} 

The outdoor dataset was collected on Vulcano Island on sandy and rocky terrain, as shown in Figure~\ref{fig:dataset}. It consists of 18 trajectories of trotting and adaptive walking, spanning over 530 m. The position ground truth was obtained using an Emlid GNSS receiver mounted on the robot. To achieve precise localization within cm-level accuracy, the receiver used Real-Time Kinematic (RTK) corrections via Long Range (LoRa) transmission from a base station accurately localized with respect to a local NTRIP caster. Ground truth was acquired at 10 Hz, whereas IMU and joint encoders were recorded at 1000 Hz. Since the robot does not have a magnetometer to measure its heading, we used the Kabsch algorithm to obtain the robot's yaw orientation from the first 2 m of each trajectory in order to align the state estimation with the ground truth coordinate frame.

\subsection{Parameter Tuning}

\subsubsection{Fixed-tuned InEKF optimization}
\label{sec:fixed-tuned}

The filter noise parameters, used as the initial values for both the $\mathbf{Q}$ and $\mathbf{R}$ adaptation schemes, as well as for the fixed-tuned InEKF, were initially handcrafted. To reduce errors due to manual tuning, we employ the \textit{gold-standard} tuning method described in \cite{greenberg2023} to optimize the initial noise parameters $\boldsymbol{\sigma}_g$ and $\boldsymbol{\sigma}_a$ corresponding to the robot orientation and linear velocity in $\mathbf{Q}$, as well as the foot measurement noise $\boldsymbol{\sigma}_l$ in $\mathbf{R}$. Given ground-truth data $( \mathbf x_t,  \mathbf z_t)$ from motion capture tracking, the $\mathbf{Q}$ and $\mathbf{R}$ matrices are obtained as follows:
\begin{align}
	\mathbf Q_t =& \operatorname{Cov}\left( \mathbf x_{t+1} - \mathbf{A}_{t}  \mathbf x_{t} \right) \\
	\mathbf R_t =& \operatorname{Cov}\left( \mathbf z_{t} - \mathbf{H}_{t}  \mathbf x_{t} \right)
\end{align} 
\noindent where $ \mathbf x_t$ is the state and $ \mathbf z_t$ is the measurement. More precisely, the prediction step of the Kalman filter is used to estimate the process noise covariance $\mathbf{Q}$. The ground-truth state at time $t$ is propagated through the robot motion model using the measured IMU angular velocity and linear acceleration. The difference between the propagated and subsequent ground-truth states is used to estimate the process noise standard deviations $\boldsymbol{\sigma}_{g}$ and $\boldsymbol{\sigma}_{a}$. The measurement noise covariance $\mathbf{R}$ is estimated from the update step using leg odometry measurements. During foot-contact phases, the relative robot displacement obtained from motion capture is compared with the corresponding displacement estimated from leg kinematics. The resulting errors are used to estimate the foot measurement noise standard deviation $\boldsymbol{\sigma}_{l}$.

The optimization is performed using indoor random trotting and adaptive trajectories \textit{L1} and \textit{L5}. The optimized noise parameters are subsequently transferred to the outdoor dataset, as precise yaw alignment between the GPS frame and the robot world frame is unavailable. This evaluates the ability of the optimized parameters to generalize to unseen environments without additional tuning. While effective for baseline tuning, this approach can still be subject to motion capture noise, numerical velocity estimation, and the linear KF assumptions underlying the \textit{gold-standard} method, which do not fully capture the nonlinear dynamics of the InEKF.

\subsubsection{Window size for innovation and residual} 
The window size plays a role in the accuracy of the innovation or residual, since averaging over shorter or longer time periods can lead to different results during dynamic robot motions. Furthermore, window sizes that are too small may lead to unstable estimation~\cite{ding2007}. Empirically, we found an optimal window size of 1 second. By defining the window in seconds rather than samples, consistency is ensured across filters running at different estimation frequencies.

\subsubsection{Scale factors for baseline approach} The hyperparameters for the baseline adaptation method based on IMU and foot force sensor data were selected using grid search on the indoor dataset as follows: $\alpha_1 = 0.1$, $\alpha_2=0.1$ and $\alpha_3=10$.

\subsection{Evaluation Metrics}

\begin{figure}[]
	\centering
	
	\subfloat[$\mathbf{Q}$ and $\mathbf{R}$ adaptation results for the baseline approach based on IMU and foot force data, compared to the proposed approach based on filter innovation and residual.]{
		\includegraphics[width=0.47\columnwidth]{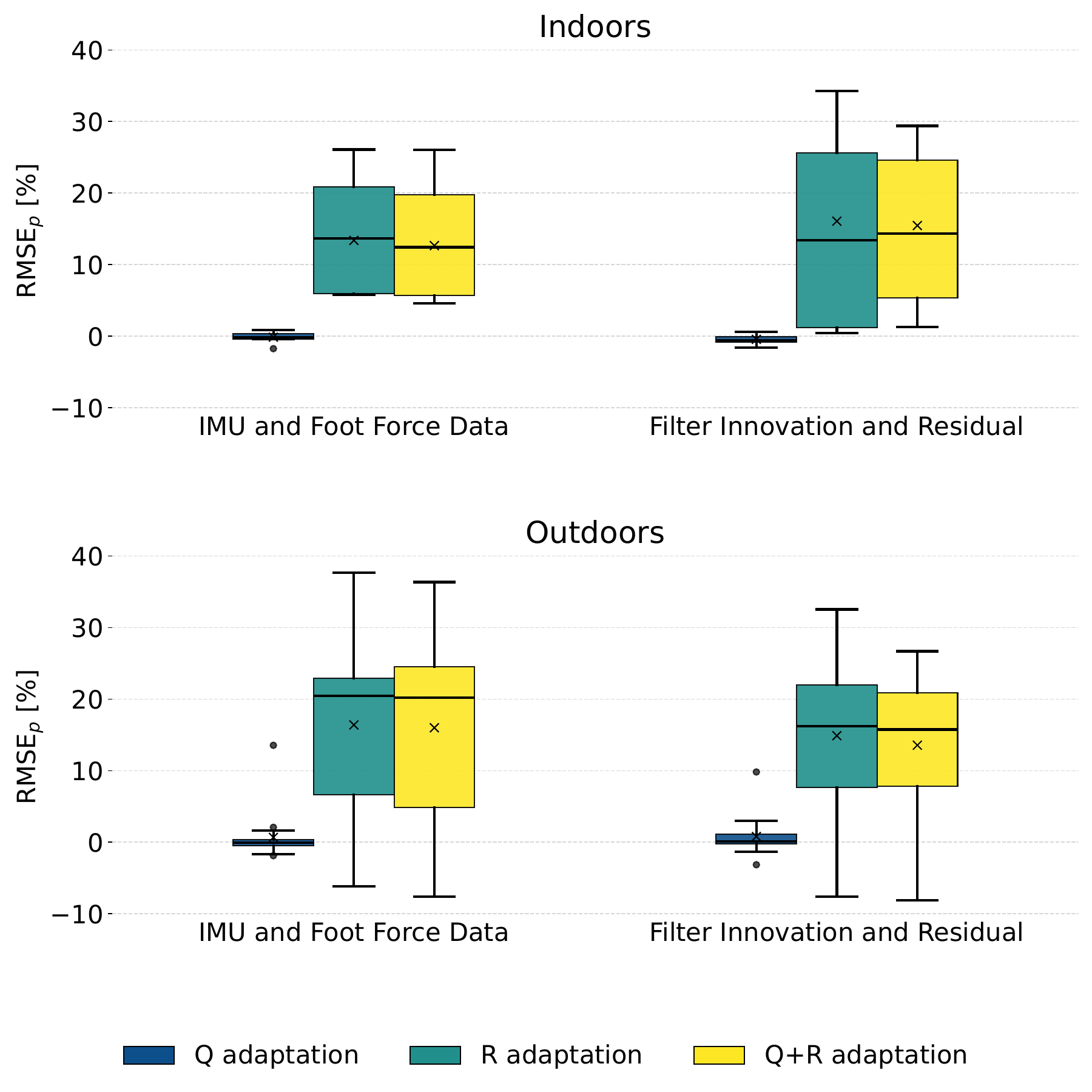}
		\label{fig:results-boxplot}
	}
	\hfill
	\subfloat[$\mathbf{R}$ adaptation results on indoor and outdoor datasets split by gait type, i.e. trotting and adaptive.]{
		\includegraphics[width=0.47\columnwidth]{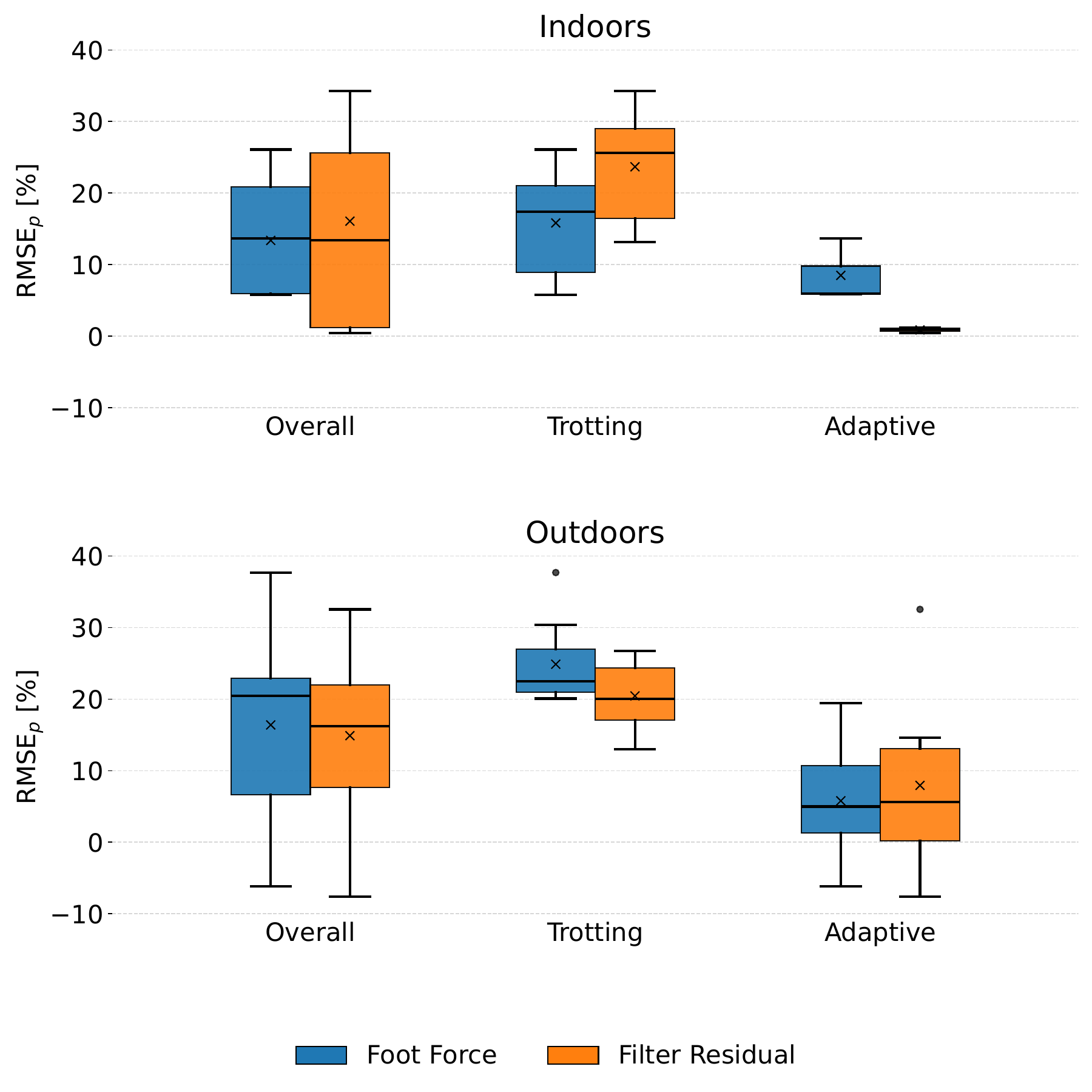}
		\label{fig:r-results-boxplot-gaits}
	}
	
	\caption{Comparison of covariance adaptation strategies for quadruped state estimation.}
	\label{fig:cov-adaptation-comparison}
\end{figure}

For evaluation, we introduce the following metrics. Firstly, the improvement in the Euclidean position root mean squared error (RMSE) for both adaptation methods relative to the fixed-parameter InEKF is defined as:
\begin{equation}
	\mathrm{RMSE}_p = \frac{\mathrm{RMSE}_{\text{fixed-tuned InEKF}} - \mathrm{RMSE}_{\text{adaptation method}}}{\mathrm{RMSE}_{\text{fixed-tuned InEKF}}} \cdot 100
\end{equation}
where higher values denote better performance. Results in \autoref{fig:cov-adaptation-comparison} are illustrated using box plots of $\mathrm{RMSE}_p$, where the mean is denoted by '$\scriptstyle \times$' and the median by the central horizontal line.

Moreover, we define the drift per distance traveled (DDT) as the difference between the final estimated and ground truth positions, normalized by the total trajectory length $D$ as follows:
\begin{equation}
	\mathrm{DDT}_i =
	\frac{\left| \hat{p}_{f,i} - p_{f,i} \right|}{D},
	\qquad i \in \{x,y,z\},
\end{equation}
where $\hat{p}_{f,i}$ and $p_{f,i}$ denote the estimated and ground-truth final robot positions along each axis $i$, respectively.

\section{Results}
\label{sec:results}

\subsection{Joint \textbf{Q} and \textbf{R} Adaptation}
\label{sec:results-jointly}

In this section, we conduct an ablation study to evaluate the impact of adapting the $\mathbf{Q}$ and $\mathbf{R}$ covariance matrices on the state estimation of legged robots, and to determine which component has the greatest influence on performance. As shown in Figure~\ref{fig:results-boxplot}, adapting $\mathbf{Q}$ alone has a negligible effect in the indoor and outdoor scenarios for both adaptation methods. In contrast, adapting $\mathbf{R}$ consistently outperforms the fixed-tuned InEKF, achieving a mean $\mathrm{RMSE}_p$ reduction of approximately 15\%. The median $\mathrm{RMSE}_p$ is slightly higher for the outdoor dataset, which we attribute to increased variability in contact conditions and terrain irregularities, as well as the fact that the initial noise parameters were tuned on the indoor dataset. Finally, jointly adapting $\mathbf{Q}$ and $\mathbf{R}$ yields performance comparable to adapting $\mathbf{R}$ alone.

Since the Kalman gain depends on the relative weighting between $\mathbf{Q}$ and $\mathbf{R}$, adapting either one is sufficient. In the following, we focus solely on adapting $\mathbf{R}$, which yields consistent improvements across all evaluated scenarios.

\subsection{Only \textbf{R} Adaptation}

In this section, we provide a detailed evaluation of the $\mathbf{R}$ matrix adaptation using the proposed filter residual method and compare it to the foot force baseline method. An overview of the obtained results is presented in \autoref{table:results-r}, where the evaluation is divided by dataset and gait type. The filter residual method performs better in terms of $\mathrm{RMSE}_p$ on the indoor dataset with a flat floor, while foot force adaptation performs slightly better on the outdoor dataset with rocky and irregular terrain. Moreover, the drift per distance traveled (DDT) shows that the filter residual-based adaptation achieves the largest reduction in drift across both datasets, with the exception of the \textit{z}-axis in the indoor scenario.

\begin{table*}[]
	\vspace{2mm}
	\centering
	\caption{Overview of state estimation results for the $\mathbf{R}$ covariance adaptation, categorized by dataset and gait. The best result for the overall trajectories in each dataset is highlighted in bold. RMSE$_{\mathrm{p}}$ represents the position RMSE improvement compared to the fixed-tuned InEKF, $\sigma_{\mathrm{p}}$ is the standard deviation of RMSE$_{\mathrm{p}}$, and max(RMSE$_{\mathrm{p}}$) is the maximum trajectory improvement. RMSE$_{\mathrm{yaw}}$ is the yaw orientation error, and DDT stands for the drift per distance traveled on each axis.}
	
	\resizebox{\textwidth}{!}{					
		\begin{tabular}{cccS[table-format=2.2]S[table-format=2.2]S[table-format=2.2]S[table-format=1.2]S[table-format=1.4]S[table-format=1.4]S[table-format=1.4]}
		\toprule
		Dataset & Gait & Method & {RMSE$_{\mathrm{p}}$ [\%] $\uparrow$} & {$\sigma_{\mathrm{p}}$ [\%] $\downarrow$} & {max(RMSE$_{\mathrm{p}}$) [\%] $\uparrow$} & {RMSE$_{\mathrm{yaw}}$ [rad] $\downarrow$} & {DDT$_x$ $\downarrow$} & {DDT$_y$ $\downarrow$} & {DDT$_z$ $\downarrow$} \\
		\midrule
		\multirow{9}{*}{Indoors} & \multirow{3}{*}{Overall} & Fixed-tuned InEKF & {\text{-}} & {\text{-}} & {\text{-}} & 0.77 & 0.0382 & 0.0261 & 0.0168 \\
		&  & Foot Force & 13.39 & \textbf{7.78} & 26.09 & 0.71 & 0.0356 & 0.0313 & \textbf{0.0134} \\
		&  & Filter Residual & \textbf{16.08} & 13.31 & \textbf{34.26} & \textbf{0.64} & \textbf{0.0323} & \textbf{0.0197} & 0.0176 \\
		\cmidrule{2-10}
		& \multirow{3}{*}{Trotting} & Fixed-tuned InEKF & {\text{-}} & {\text{-}} & {\text{-}} & 0.70 & 0.0287 & 0.0284 & 0.0225 \\
		&  & Foot Force & 15.83 & 8.20 & 26.09 & 0.68 & 0.0284 & 0.0346 & 0.0191 \\
		&  & Filter Residual & 23.68 & 8.69 & 34.26 & 0.46 & 0.0186 & 0.0177 & 0.0233 \\
		\cmidrule{2-10}
		& \multirow{3}{*}{Adaptive} & Fixed-tuned InEKF & {\text{-}} & {\text{-}} & {\text{-}} & 0.92 & 0.0572 & 0.0217 & 0.0052 \\
		&  & Foot Force & 8.51 & 4.49 & 13.69 & 0.78 & 0.0499 & 0.0246 & 0.0019 \\
		&  & Filter Residual & 0.88 & 0.41 & 1.21 & 1.00 & 0.0597 & 0.0238 & 0.0062 \\
		\midrule
		\multirow{9}{*}{Outdoors} & \multirow{3}{*}{Overall} & Fixed-tuned InEKF & {\text{-}} & {\text{-}} & {\text{-}} & {\text{-}} & 0.0837 & 0.0851 & 0.0514 \\
		&  & Foot Force & \textbf{16.39} & 11.86 & \textbf{37.68} & {\text{-}} & 0.0807 & 0.0835 & 0.0447 \\
		&  & Filter Residual & 14.89 & \textbf{10.68} & 32.54 & {\text{-}} & \textbf{0.0805} & \textbf{0.0807} & \textbf{0.0435} \\
		\cmidrule{2-10}
		& \multirow{3}{*}{Trotting} & Fixed-tuned InEKF & {\text{-}} & {\text{-}} & {\text{-}} & {\text{-}} & 0.0408 & 0.0741 & 0.0782 \\
		&  & Foot Force & 24.87 & 5.62 & 37.68 & {\text{-}} & 0.0403 & 0.0723 & 0.0629 \\
		&  & Filter Residual & 20.44 & 4.73 & 26.74 & {\text{-}} & 0.0409 & 0.0753 & 0.0662 \\
		\cmidrule{2-10}
		& \multirow{3}{*}{Adaptive} & Fixed-tuned InEKF & {\text{-}} & {\text{-}} & {\text{-}} & {\text{-}} & 0.1373 & 0.0988 & 0.0178 \\
		&  & Foot Force & 5.79 & 8.35 & 19.46 & {\text{-}} & 0.1312 & 0.0976 & 0.0220 \\
		&  & Filter Residual & 7.95 & 12.22 & 32.54 & {\text{-}} & 0.1301 & 0.0874 & 0.0152 \\
		\bottomrule
	\end{tabular}}
	\label{table:results-r}
\end{table*}

Furthermore, we investigate the spread of $\mathrm{RMSE}_p$, namely the standard deviation $\sigma_p$ on two gaits, trotting and adaptive, as shown in Figure~\ref{fig:r-results-boxplot-gaits}. On average, adaptation improves the trotting gait results by up to 25\%, whereas it only improves the adaptive gait results by around 8\%, possibly due to firmer and more regular ground contact during the trotting gait. Additionally, the proposed residual adaptation method outperforms the baseline foot force approach on the indoor trotting and outdoor adaptive gaits.

\begin{figure}[]
	\centering
	
	\subfloat[Indoors \textit{L7}, top view.]{%
		\includegraphics[width=0.24\columnwidth]{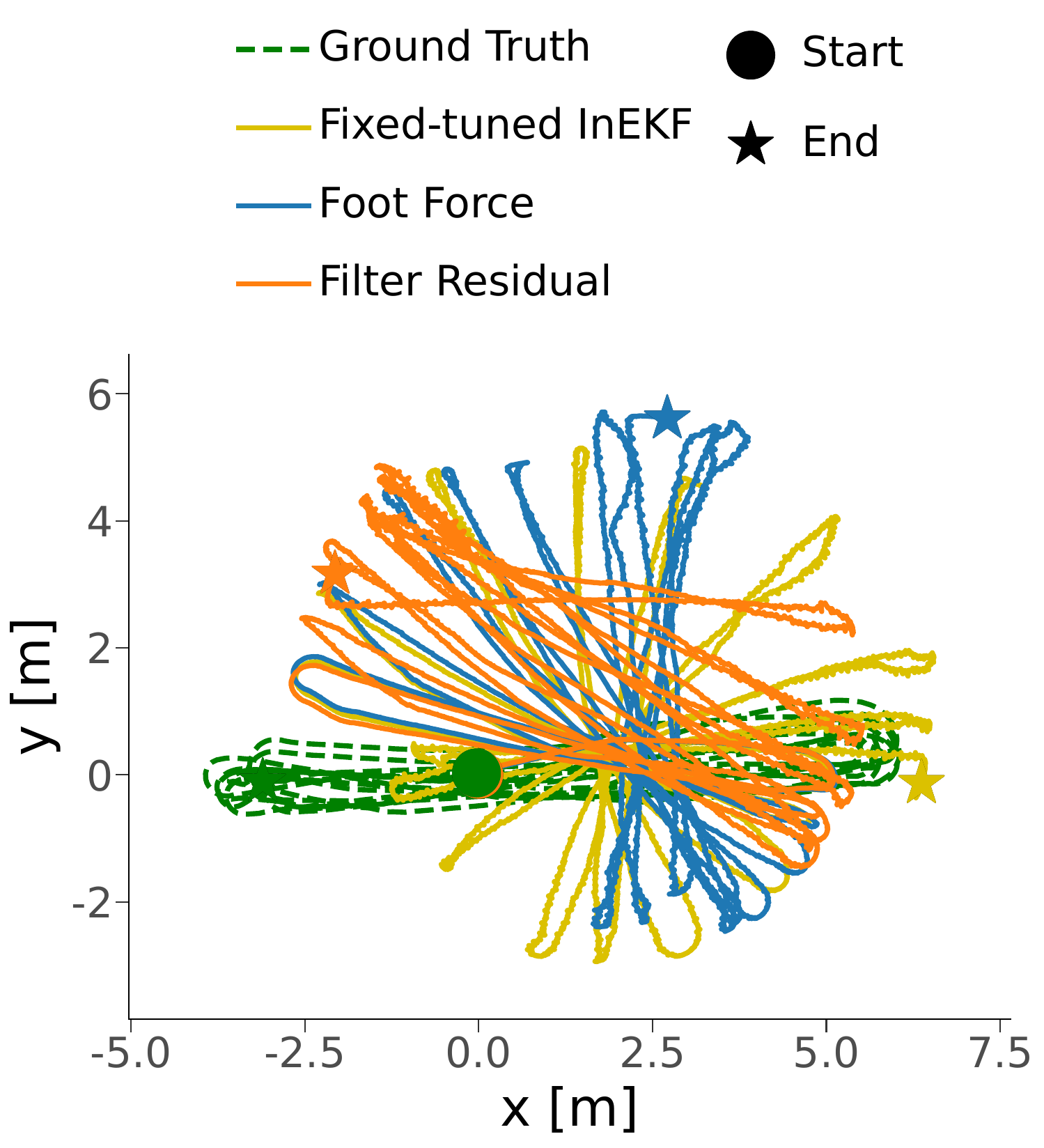}
		\label{fig:xy-plane-lab}%
	}
	\hfill
	\subfloat[Indoors \textit{L7}, 3-axis view.]{%
		\includegraphics[width=0.225\columnwidth]{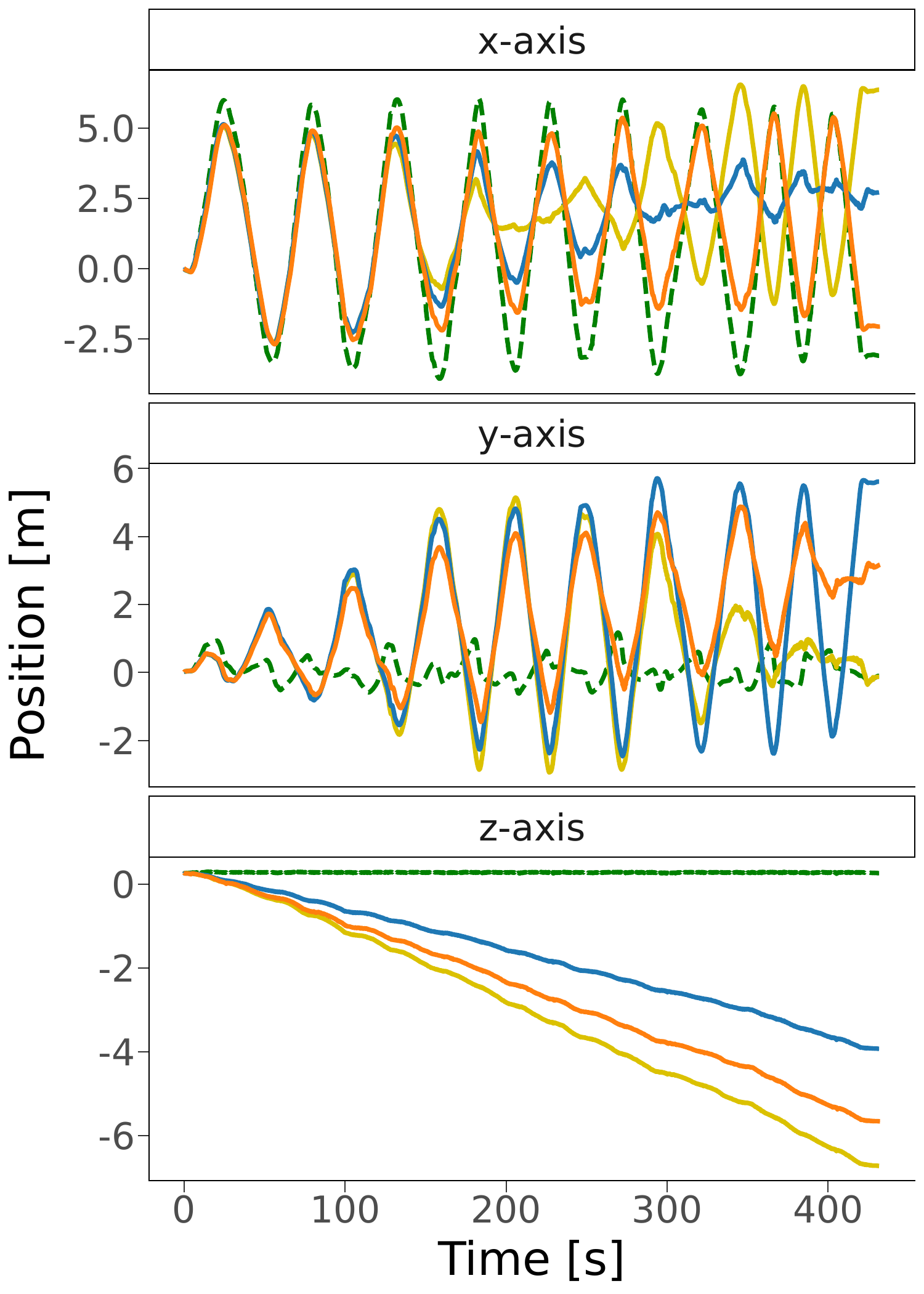}
		\label{fig:xyz-plot-lab}%
	}
	\hfill
	\subfloat[Outdoors \textit{F10}, top view.]{%
		\includegraphics[width=0.215\columnwidth]{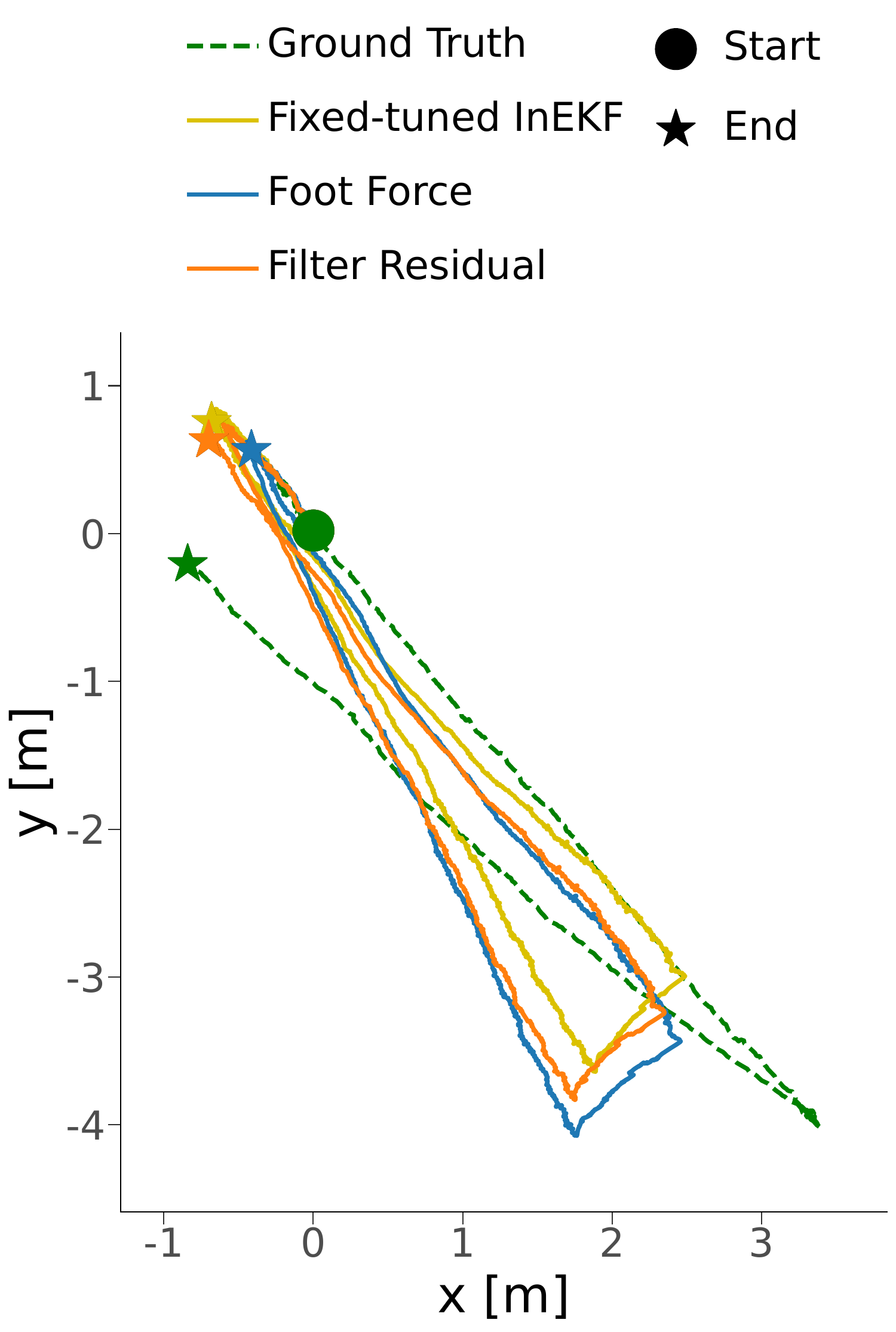}
		\label{fig:xy-plane-field}%
	}
	\hfill
	\subfloat[Outdoors \textit{F10}, 3-axis view.]{%
		\includegraphics[width=0.225\columnwidth]{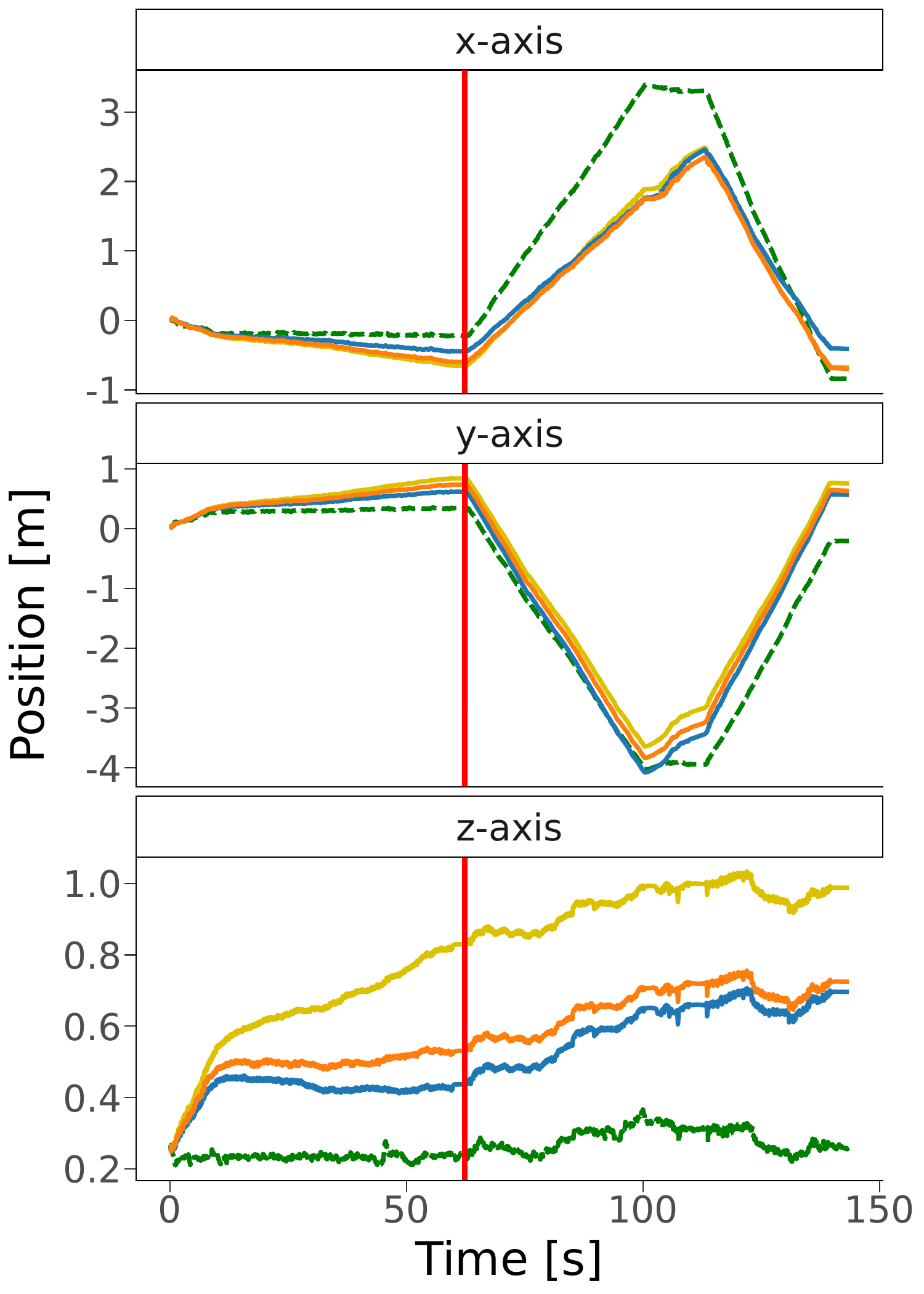}
		\label{fig:xyz-plot-field}%
	}
	
	\caption{Indoor and outdoor trajectories with $\mathbf{R}$ matrix adaptation. Indoor RMSE: 5.97~m (fixed-tuned InEKF), 4.41~m (foot force), 4.17~m (filter residual). Outdoor RMSE: 1.05~m (fixed-tuned InEKF), 0.80~m (foot force), 0.91~m (filter residual). The vertical red line in \textit{F10} indicates the transition from stepping in place to trotting.}	
	\label{fig:trajectories-r-adaptation}
\end{figure}

\begin{figure}[]
	\centering
	\includegraphics[width=\linewidth]{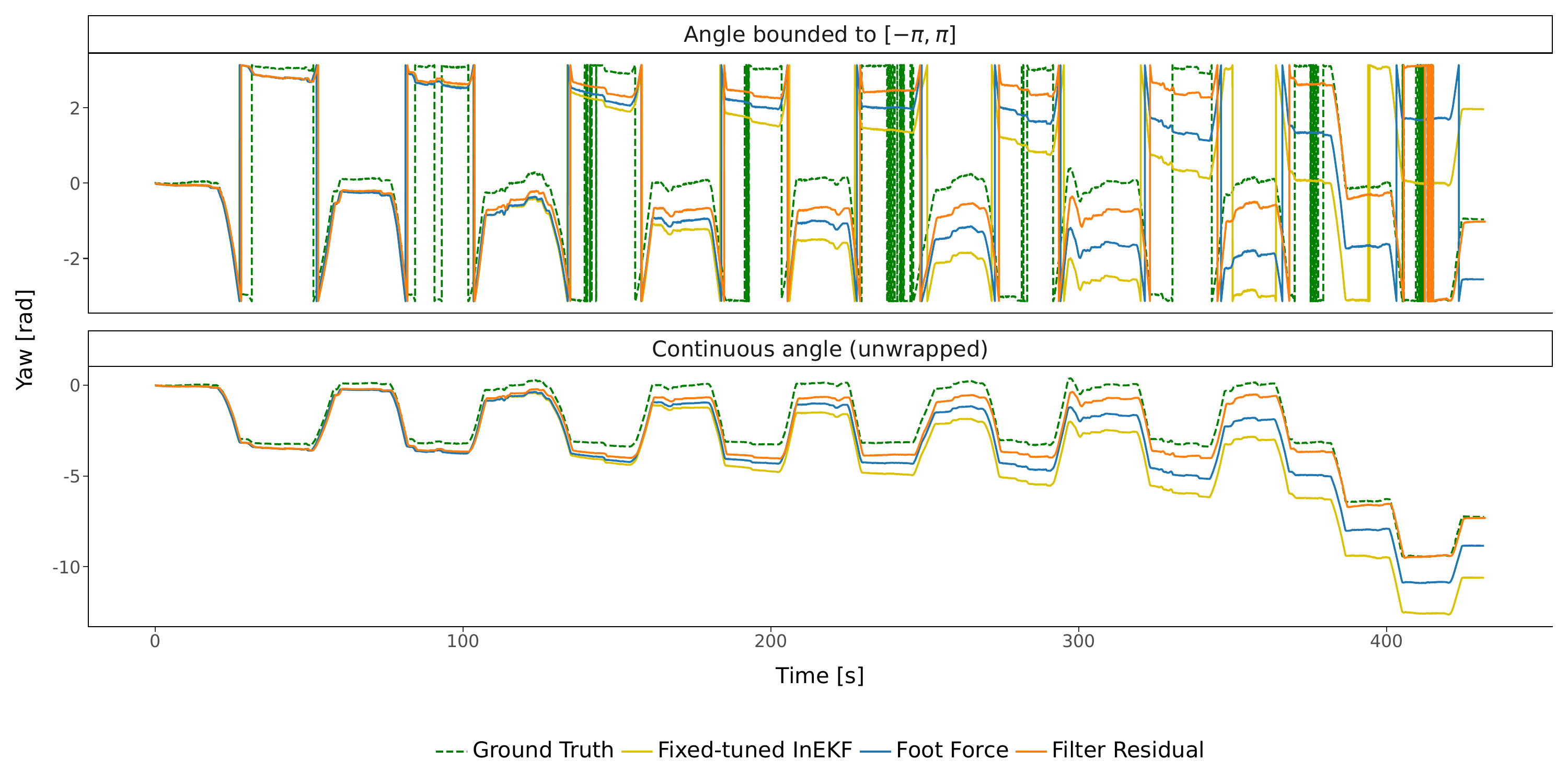}
	\caption{Yaw angle estimation for indoor trajectory \textit{L7} with $\mathbf{R}$ covariance matrix adaptation. The top subplot shows the bounded representation in the range $[-\pi,\pi]$, while the bottom subplot displays the continuous unwrapped yaw evolution. The robot follows a straight-line trajectory and performs 180-degree turns at the boundaries of the Vicon tracking area.}		
	\label{fig:rpy-plot-lab}
\end{figure}

Next, \autoref{fig:trajectories-r-adaptation} presents the \textit{L7} trajectory from the indoor dataset and the \textit{F10} trajectory from the outdoor dataset. Adaptation significantly reduces drift along the $z$-axis and improves the heading estimate, resulting in better alignment of the \textit{L7} trajectory with the ground truth under filter residual adaptation.
Since yaw errors can accumulate over time during turning motions, leading to increased positional drift, accurate yaw estimation plays a critical role in achieving overall accuracy. \autoref{fig:rpy-plot-lab} shows the yaw evolution for the \textit{L7} trajectory, where the yaw angle is estimated most accurately using the proposed filter residual method. The corresponding outdoor yaw results for \textit{F10} are omitted, as GPS-based ground truth does not provide orientation data. A complete overview of the yaw RMSE for the indoor dataset is presented in \autoref{table:results-r}, where residual-based adaptation yields the largest overall improvement.

\begin{figure}[!htbp]
	\centering
	
	\subfloat[Indoor trotting (\textit{L7})]{%
		\includegraphics[width=0.45\columnwidth]{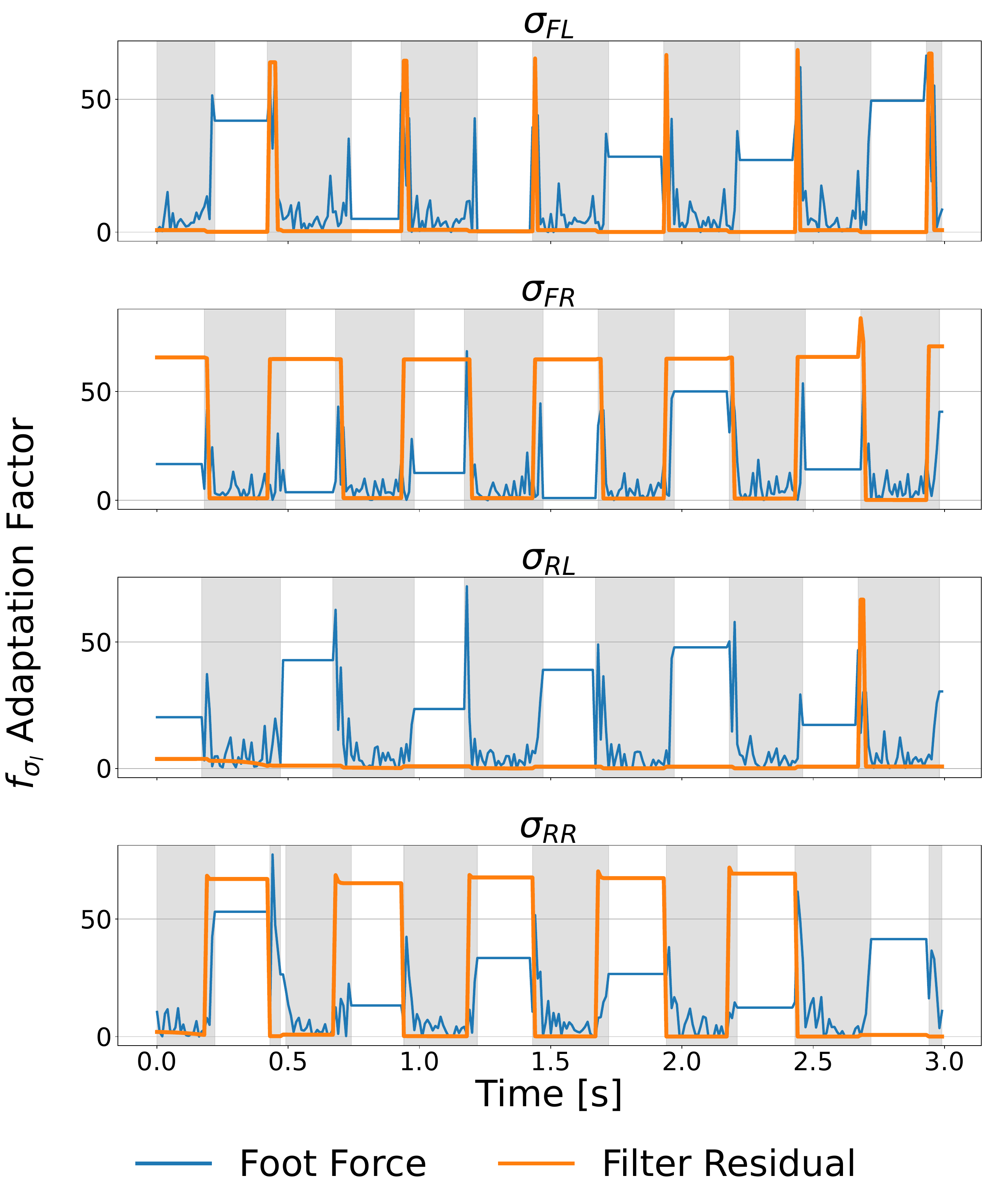}
		\label{fig:r-formula-adaptation-trot-lab}%
	}
	\hfill
	\subfloat[Outdoor trotting (\textit{F10})]{%
		\includegraphics[width=0.45\columnwidth]{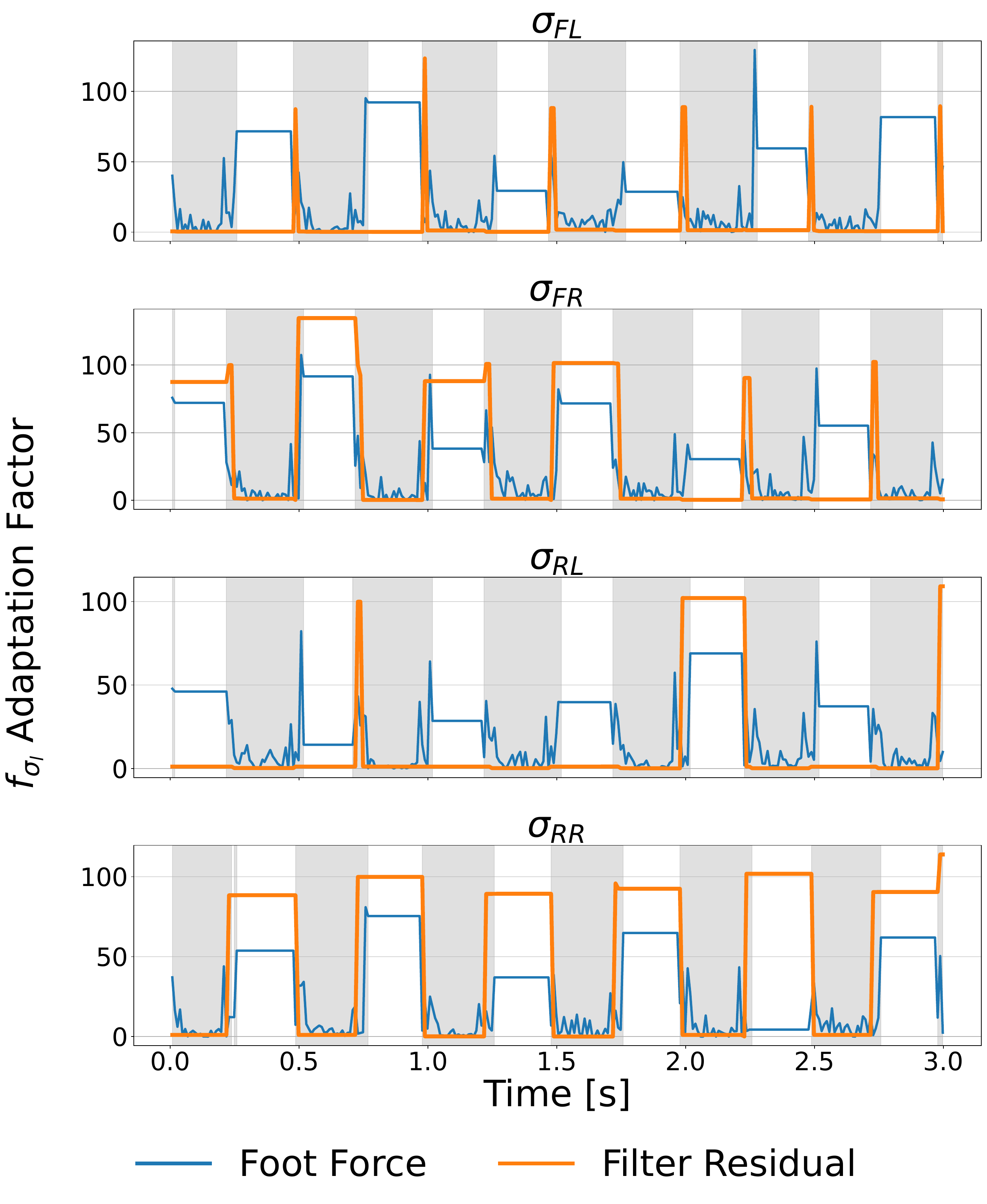}
		\label{fig:r-formula-adaptation-trot-field}%
	}
	
	\subfloat[Indoor adaptive trot (\textit{L5})]{%
		\includegraphics[width=0.45\columnwidth]{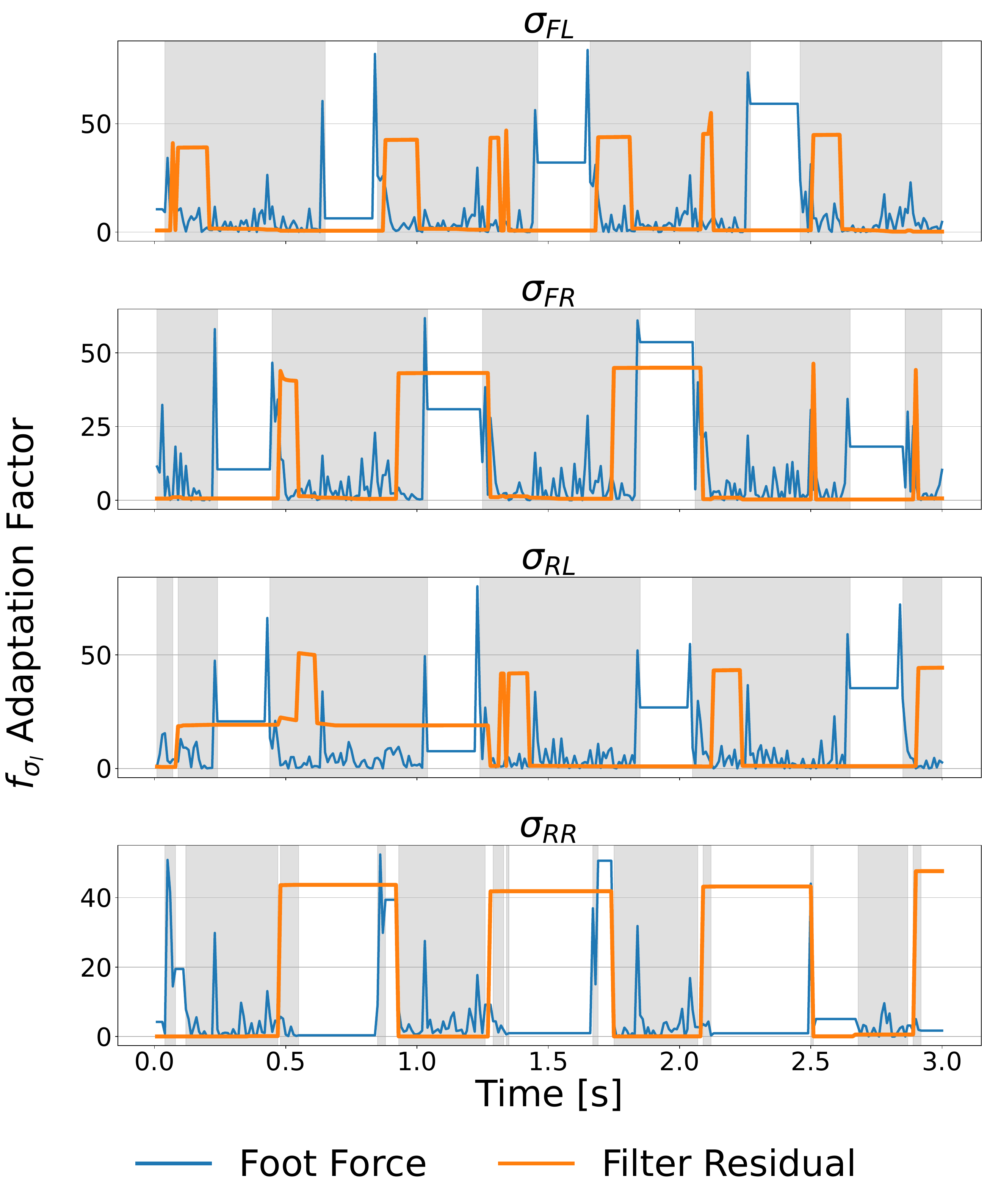}
		\label{fig:r-formula-adaptation-adaptive-lab}%
	}
	\hfill
	\subfloat[Outdoor adaptive trot (\textit{F10})]{%
		\includegraphics[width=0.45\columnwidth]{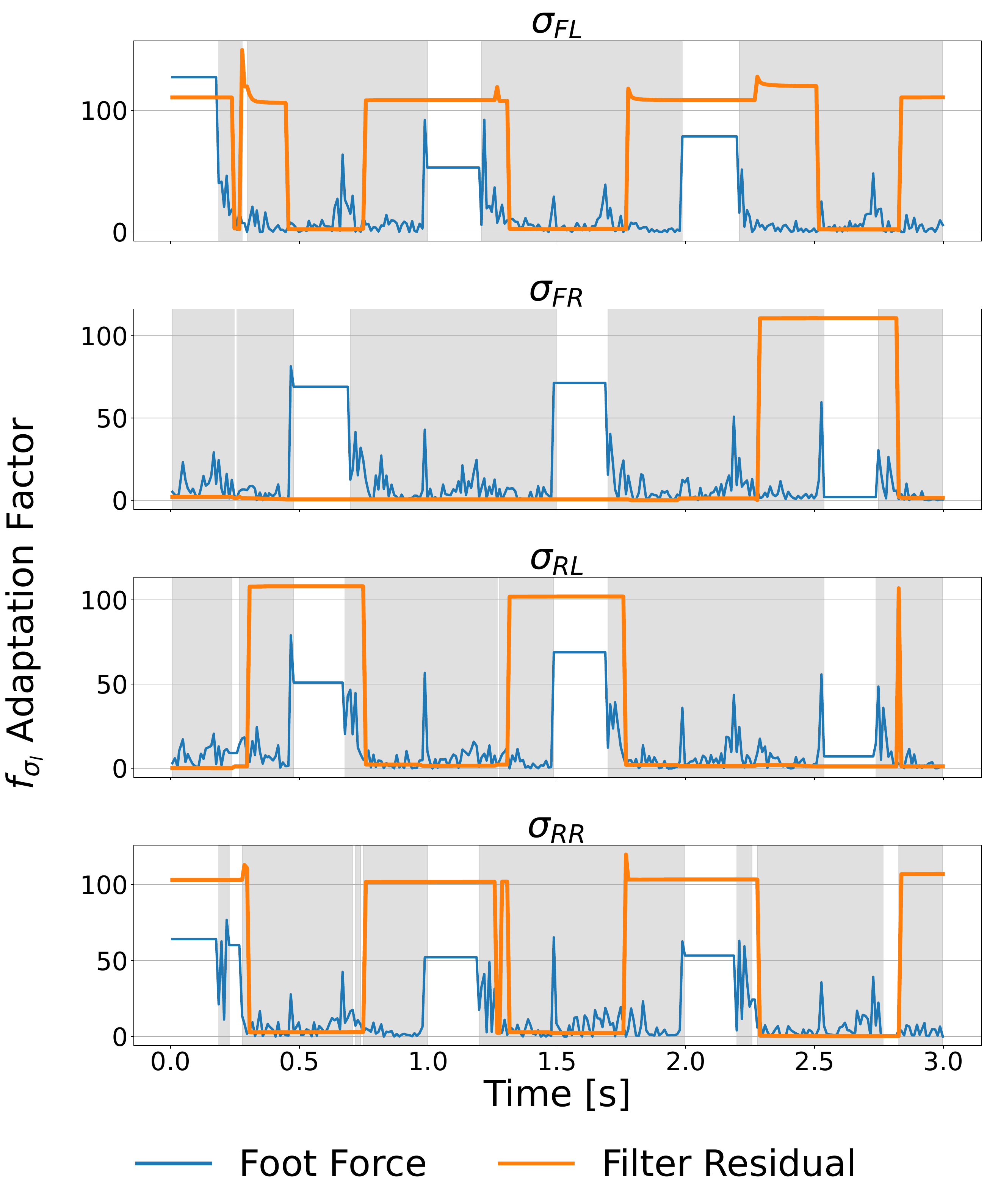}
		\label{fig:r-formula-adaptation-adaptive-field}%
	}
	
	\caption{The $\mathbf{R}$ adaptation factor for both methods is plotted over 3 s for each foot (FL, FR, RL and RR) for trotting and adaptive gaits. The contact sequences are highlighted in gray when the corresponding foot is in contact with the ground.}
	\label{fig:r-formula-adaptation}
\end{figure}

Finally, \autoref{fig:r-formula-adaptation} illustrates the $\mathbf{R}$ matrix adaptation factor defined as  $f_{\sigma_l} = |\sigma_l - \sigma_{l_0}|/\sigma_{l_0}$ overlapped on the gait sequence, where the gray regions indicate ground contact phases. For both methods, the value of $\sigma_l$ for each foot, namely $\left[\sigma_{FL}, \sigma_{FR}, \sigma_{RL}, \sigma_{RR}\right]$, is updated only during ground contact and remains constant otherwise. Although the foot-force-based adaptation exhibits slightly more regular patterns, the residual-based method follows a comparable trend and it effectively captures the contact-driven structure triggered by leg kinematic measurements.

\section{Discussion}
\label{sec:discussion}

In the ablation study from Section~\ref{sec:results-jointly}, we found that adapting $\mathbf{R}$ has a significantly larger impact on estimation performance than adapting $\mathbf{Q}$, whether applied independently or jointly. This suggests that quadruped state estimation performance is more sensitive to the modeling of measurement uncertainty than to the process noise. This can be attributed to the fact that leg odometry, used to correct IMU-driven state propagation, exhibits strong time-varying uncertainty due to contact conditions, slip, and terrain variability. Consequently, accurately capturing its stochastic properties through the $\mathbf{R}$ covariance matrix has a direct influence on the filter performance. While the IMU is also affected by uncertainties such as bias and integration drift, these effects are primarily addressed through state augmentation and bias estimation within the filter. Transient disturbances, such as impacts and vibrations, may also affect the IMU measurements \cite{Driessen2021, Allione2023}, however their effect is not significantly reduced by the noise covariance adaptation. 

In this work, we treat leg kinematics-related uncertainties, including foot slip, encoder noise, and geometric modeling errors, as measurement uncertainty. This is achieved by adapting the measurement noise covariance matrix $\mathbf{R}$, thereby reducing confidence in leg odometry when contact quality degrades. In contrast, \cite{kim2025} models foot slip as process uncertainty and adapts the corresponding foot linear velocity component of the process noise covariance matrix $\mathbf{Q}$. This leads to another key difference, since $\mathbf{R}$ adaptation directly adjusts the Kalman gain at the current time step, whereas adapting $\mathbf{Q}$ influences the state prediction and recursively propagates through the state covariance matrix $\mathbf{P}$. Consequently, the proposed $\mathbf{R}$ adaptation remains restricted to the current measurement update, limiting state covariance drift through recursive error propagation.

Moreover, \autoref{fig:trajectories-r-adaptation} reveals that the $z$-axis drift is negative on a flat floor indoors and positive on rocky outdoor terrain. Although contact is rigid in the indoor scenario, modeling errors and leg flexibility may introduce biases, resulting in inaccurate measurements indicating that the leg is more extended than it actually is. Conversely, on rocky terrain, the contact is less stable, leading to early contact detection before the foot is firmly placed on the ground. Consequently, the leg kinematics measurements are taken too early, leading to an uphill walking effect. Overall, the $z$-axis drift is smaller indoors, since modeling errors lead to less bias than unstable contact due to rocky terrain.

Finally, as shown in \autoref{table:results-r}, the results demonstrate the effectiveness of the proposed filter residual adaptation, particularly in controlled environments such as on a flat surface or during regular gait patterns such as trotting. Furthermore, it has the advantage of not requiring foot force sensor data, which is prone to failure or noise. However, since this method depends on the filter noise parameters, the adaptation factor may drift over time as the covariance matrix $\mathbf{P}$ steadily increases in a proprioceptive-based estimator. In practice, however, it was proven that estimation drift does not affect the residual adaptation results, even on long trajectories such as \textit{L7}.

\section{Conclusion}
\label{sec:conclusion}

We addressed the uncertainty in proprioceptive state estimation of legged robots by employing an online adaptation of the noise covariance matrices $\mathbf{Q}$ and $\mathbf{R}$ in the InEKF. Specifically, we proposed an approach based on (i) filter innovation and residual and compared it to an (ii) IMU and foot force method using real data from a quadruped robot in both indoor and outdoor environments. The results show that $\mathbf{R}$ adaptation is sufficient for improving state estimation performance. While both adaptation techniques improved performance by up to 25\% for the trotting gait compared to the fixed-tuned InEKF, the proposed residual-based method has the advantage of not requiring foot force data or parameter tuning. Finally, we released the quadruped dataset open-source, covering a total distance of 1200~m.

Future work includes validating the adaptive state estimation method in closed-loop control, as well as conducting robustness and stability tests under varying initial noise parameters. Planned gait sequences could be used to adapt the $\mathbf{R}$ covariance matrix, and the method could be evaluated on other legged platforms, such as humanoids. Additionally, an in-depth, formal analysis of the impact of joint $\mathbf{Q}$ and $\mathbf{R}$ noise adaptation could provide further insights.

\section*{Funding}

This research was done in the AAPLE project (grant 50WK2275) funded through the German Federal Ministry of Economic Affairs and Climate Action (BMWK) and \mbox{ActGPT} project (grant 01IW25002) funded by the German Aerospace Center (DLR) with federal funds from the German Ministry of Research, Technology and Space (BMFTR), with partial support from the Robotics Institute Germany (RIG).

\printcredits

\bibliographystyle{elsarticle-num}
\bibliography{cas-refs}

\end{document}